\documentclass[10pt,twocolumn,letterpaper]{article}

\usepackage{cvpr}              

\usepackage{graphicx}
\usepackage{amsmath}
\usepackage{amssymb}
\usepackage{booktabs}

\usepackage{amssymb}

\usepackage{graphicx}
\usepackage{amsmath}
\usepackage{amssymb}
\usepackage{booktabs}
\usepackage{algorithm}
\usepackage{algpseudocode}
\usepackage{multirow}
\usepackage{multicol}
\usepackage{xcolor}
\usepackage{colortbl}

\usepackage{subcaption}
\usepackage{tabularx}

\usepackage[pagebackref,breaklinks,colorlinks]{hyperref}

\usepackage[capitalize]{cleveref}
\crefname{section}{Sec.}{Secs.}
\Crefname{section}{Section}{Sections}
\Crefname{table}{Table}{Tables}
\crefname{table}{Tab.}{Tabs.}

\def\cvprPaperID{*****} 
\def\confName{CVPR}
\def\confYear{2022}

\newcommand{\red}{\textcolor{black}}
\newcommand{\redd}{\textcolor{black}}
\newcommand{\reddd}{\textcolor{red}}
\begin{document}

\title{PFB-Diff: Progressive Feature Blending Diffusion for Text-driven Image Editing}

%

\author{Wenjing Huang \quad \quad
	Shikui  Tu* \quad \quad
	Lei Xu* \\
	Shanghai Jiao Tong University
}

%

\maketitle

\begin{abstract}
			Diffusion models have demonstrated their ability to generate diverse and high-quality images, sparking considerable interest in their potential for real image editing applications. However, existing diffusion-based approaches for local image editing often suffer from undesired artifacts due to the latent-level blending of the noised target images and diffusion latent variables, which lack the necessary semantics for maintaining image consistency. To address these issues, we propose PFB-Diff, a Progressive Feature Blending method for Diffusion-based image editing. Unlike previous methods, PFB-Diff seamlessly integrates text-guided generated content into the target image through multi-level feature blending. The rich semantics encoded in deep features and the progressive blending scheme from high to low levels ensure semantic coherence and high quality in edited images. Additionally, we introduce an attention masking mechanism in the cross-attention layers to confine the impact of specific words to desired regions, further improving the performance of background editing \red{and multi-object replacement}. PFB-Diff can effectively address various editing tasks, including object/background replacement and object attribute editing. Our method demonstrates its superior performance in terms of editing accuracy and image quality without the need for fine-tuning or training. Our implementation is
available at \url{https://github.com/CMACH508/PFB-Diff}.
\end{abstract}

\section{Introduction}

\begin{figure*}[t]
	\centering
	\captionsetup{type=figure}
	\includegraphics[width=\textwidth]{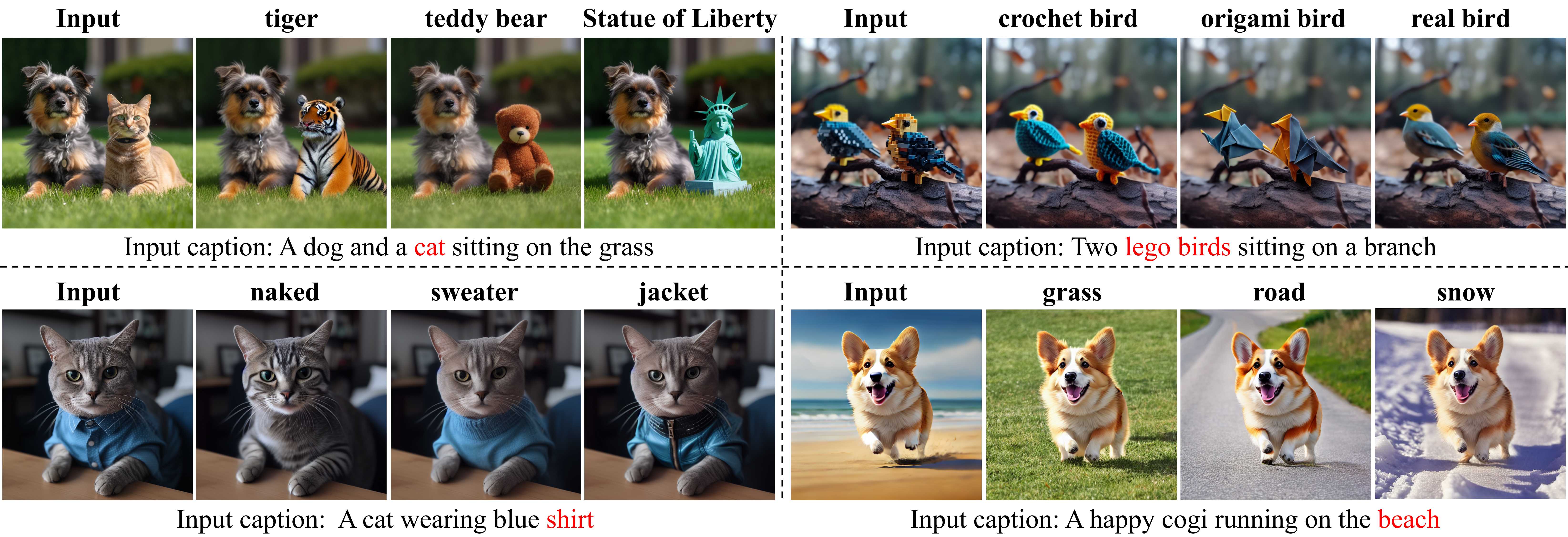}
	\captionof{figure}{PFB-Diff for real image editing. Given an image, a text query, and a coarse mask that annotates the regions of interest, our method enables text-driven image editing while preserving the irrelevant regions. Our approach can be applied to various tasks, such as object replacement (top-left), secondary object editing (bottom-left), object attribute editing (top-right), and background replacement (bottom-right). Better viewed online in color and zoomed in for details.}
	\label{fig:teaser}
\end{figure*}

Diffusion Models (DM) \cite{ddpm,guided-diff} have garnered increasing attention due to their powerful ability to synthesize high-quality and diverse images based on text prompts. Recently introduced large-scale training of diffusion models, such as Imagen \cite{imagen}, GLIDE \cite{glide}, DALL$\cdot$E3 \cite{dalle3}, Emu Video \cite{emuvideo}, and Stable Diffusion \cite{ldm}, have further improved the quality and demonstrated unprecedented semantic generation. Large-scale text-to-image models are trained on enormous amounts of image-caption pairs, resulting in a high capacity for both textual semantic understanding and image synthesis. Given a corresponding text description, they can generate realistic images that match the text description. Such success inspires subsequent efforts to leverage large-scale pre-trained diffusion models for real image editing. 

However, extending pre-trained text-to-image frameworks for personalized editing (e.g., manipulating a specific object in an image) remains a challenging task due to the limited ability to describe desired objects through text. In fact, even the most comprehensive and detailed textual description of an object may result in instances of different appearances.
Besides, the editing process should meet the requirements of high accuracy (or image-text alignment), image consistency, irrelevance preservation, and image fidelity. Image consistency measures whether newly generated content in a target region exhibits contextual coherence both semantically and textually. For instance, when adding a teddy bear to a tree branch, it should align with the environment, such as adopting a tree climbing pose (semantic consistency), and seamlessly blend with the original image (texture consistency).

To achieve personalized editing, some pioneering works \cite{textual-inversion, imagic, dreambooth} proposed to learn a unique word to represent the given object and fine-tune pre-trained text-to-image diffusion models on several images containing the same object. By inserting its unique word in various text contexts, they can synthesize the object in diverse scenes and manipulate it with text guidance. These methods can typically generate high-fidelity images, but sometimes, they drastically alter the content of the original image. Besides, these methods cannot fully leverage the generalization ability of the pre-trained model due to the over-fitting issues \cite{sine} or language drift problem \cite{dreambooth}, i.e., the model gradually loses the ability to generate subjects that belong to the same category as the target object. Moreover, the above methods introduce a per-image optimization process, which cannot satisfy the requirement of high efficiency in practical applications.

In contrast, the optimization-free methods \cite{diffedit,bdm,bldm,sdedit} perform image editing based on a frozen pre-trained diffusion model and require no optimization or fine-tuning. To seamlessly fuse newly generated content with the original image, they generally apply the diffusion process first on the input image and then conduct the denoising procedure conditioned on both the input image and the text guidance. 
At each time step of the denoising process, the noisy versions of the input image are typically blended with the text-guided diffusion latent variables using either a user-provided mask \cite{bdm,bldm,sdedit} or a self-predicted mask \cite{diffedit}. Without fine-tuning or optimization, the above methods can avoid the language drift problem \cite{dreambooth} and run in real-time.
Additionally, employing the original image and mask for iterative refinement can aid in preserving visual details.

\begin{figure*}[t]
	\centering
	\includegraphics[width=\textwidth]{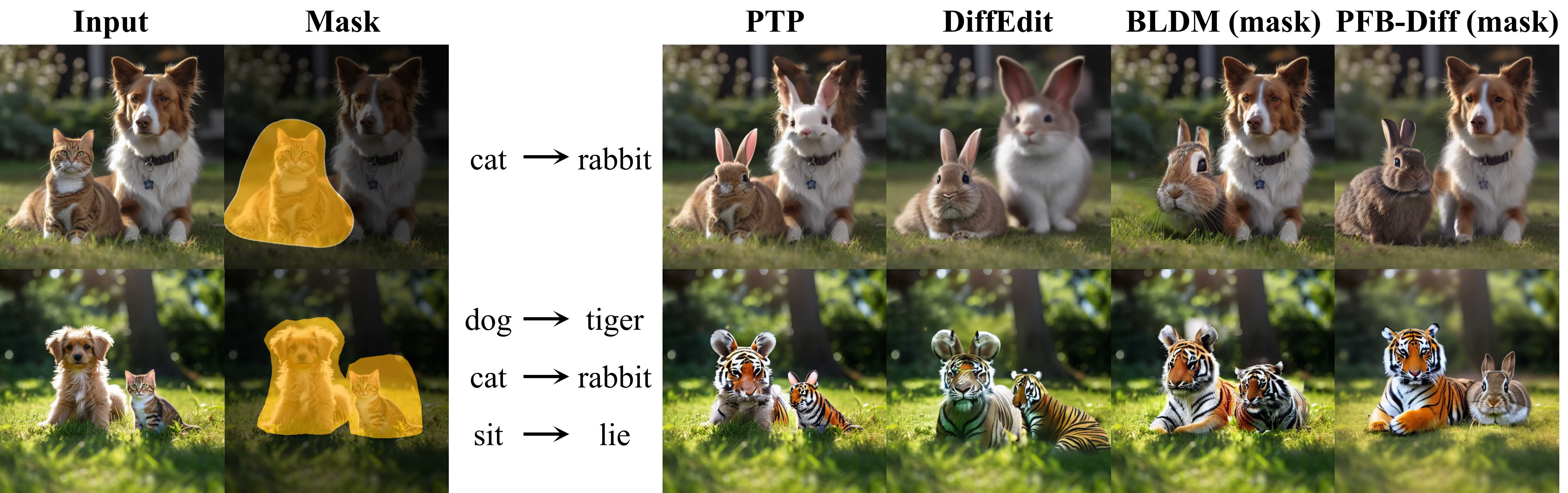}
	\caption{\red{Examples of text-driven image edits. The second column displays the rough masks provided to BLDM\cite{bldm} and PFB-Diff. In the first row, DiffEdit \cite{diffedit} and PTP \cite{ptp} can cause undesired modifications to the dog, while BLDM\cite{bldm} cannot generate a complete rabbit. In the second row, traditional methods are unable to accurately implement object replacement in multi-object scenarios. PFB-Diff can achieve accurate editing while maintaining high image quality.}}
	\label{fig:introduction}
\end{figure*}

However, most optimization-free methods face several challenges, as illustrated in Figure \ref{fig:introduction}. Firstly, many methods rely on editing intermediate noisy images \cite{diffedit, bdm, bldm, sdedit}, which lack the necessary semantics for maintaining image consistency, resulting in degraded outcomes. Additionally, some methods \cite{bldm, diffedit,sdedit} struggle to accurately generate desired content at specified locations based on textual prompts. \red{ Additionally, most existing methods focus on editing a single object. When multiple objects need to be replaced simultaneously, the performance of these methods often falls short of expectations.}

\red{Among optimization-free methods, mask-free image editing techniques such as PTP \cite{ptp} and DiffEdit\cite{diffedit} automatically predict masks based on input text prompts, eliminating the need for user-provided masks. While this approach is more convenient for users, the predicted editing areas often do not align with the user's areas of interest, particularly in images containing multiple objects.}
In commercial applications like Midjourney\footnote{\url{https://www.midjourney.com/home/?callbackUrl=\%2Fapp\%2F}} and open-source projects such as Stable Diffusion web UI\footnote{\url{https://github.com/AUTOMATIC1111/stable-diffusion-webui}}, it is common for users to provide rough masks. Additionally, recent segmentation models such as Segment Anything Model (SAM) \cite{sam} can generate high-quality object masks based on input prompts like points or boxes, simplifying the mask acquisition process for users. Therefore, this paper aims to develop a method based on rough masks. Employing a rough mask requires little user effort, yet it enhances flexibility and ensures more reliable editing results. 

In this paper, we propose PFB-Diff, a framework adapting pre-trained text-to-image models for text-driven image editing. As an optimization-free method, our method inherits its advantages of efficiency and the ability to fully utilize the pre-trained model's generalization capabilities. Meanwhile, we propose Progressive Feature Blending (PFB) to overcome the existing issues in previous optimization-free methods, such as image inconsistency and image-text misalignment. Specifically, instead of editing intermediate noisy images, we propose to edit their deep feature maps in the prediction network. Deep features contain rich semantics, allowing feature-level editing to preserve semantic and texture consistency in edited images effectively. The editing is progressively performed from high to low feature levels via multi-scale masks. This approach helps to seamlessly integrate the newly generated content into the original image, resulting in more natural and coherent synthesis results.

\red{Due to the limited capabilities of Stable Diffusion \cite{ldm} in language understanding and image generation, generating multiple objects simultaneously often leads to mutual interference, causing some objects not being accurately generated. Furthermore, the model sometimes generates additional foreground objects in the background during background replacement. To address these issues, we introduce the Attention Masking (AM) mechanism in PFB-Diff. This mechanism selectively adjusts the attention maps of specific words using user-provided masks, thereby restricting their influence to defined regions. This approach not only enhances background editing but also ensures that the model accurately generates target objects at specified locations, particularly in multi-object replacement scenarios. }

Attention editing techniques are frequently employed in concurrent works such as Prompt-to-prompt \cite{ptp} and FateZero \cite{fatezero}, which use cross-attention injections for image editing or use cross-attention layers for mask prediction. However, our approach introduces a unique variation. PFB-Diff stands out as the first method to limit the impact of a word by refining its cross-attention maps through a coarse mask provided by the user.

\textcolor{black}{To quantitatively evaluate the performance of image editing, we collect 9,843 images from the COCO dataset \cite{coco} to build the COCO-animals-10k dataset. This dataset comprises 9,843 object replacement instances and 1,597 background replacement instances. 
	Both quantitative and qualitative experiments show that our method outperforms current leading image editing methods.}

\red{To sum up, the contribution of this work lies in three folds:
	\begin{itemize}
		\item We propose a training-free method that effectively addresses various editing tasks, such as object/background replacement and object attribute editing.
		\item  To seamlessly integrate newly generated content into the original image, we propose a progressive feature blending (PFB) module. Instead of editing intermediate noisy images, the PFB blends deep feature maps from high to low levels, ensuring semantic coherence in edited images.
		\item We introduce an attention masking mechanism that adjusts the attention maps of specific words using user-provided masks, enabling the model to accurately generate target objects at specified locations.
\end{itemize}}

\red{In the following, the paper is organized as follows. Section 2 introduces related work relevant to our study. Section 3 provides background knowledge on diffusion models. Section 4 describes our proposed method in detail.  Section 5 presents the experimental results and their analysis. Section 6 discusses the limitations of our method and possible solutions. Finally, Section 7 concludes the paper.
}

\section{Related works}

\subsection{Semantic image editing}
Semantic image editing has received widespread attention from the vision and graphics community due to its various potential applications. Rapid advances in generative adversarial networks (GANs) \cite{gan} enable users to edit images using various instructions, such as attribute labels \cite{stylegan,interfacegan}, facial landmarks \cite{ganimation}, spatial masks \cite{mask-guided}, or even the text \cite{clip2stylegan}. In recent years, with the emergence of StyleGAN \cite{stylegan, stylegan2},  image generation and editing have achieved significant improvements in quality. To fully utilize the generative capabilities of StyleGAN \cite{stylegan, stylegan2}, a large number of recent works \cite{image2stylegan, image2stylegan++, ganspace, indomain,yang2021semantic,interfacegan} focus on exploiting the rich interpretable semantics in the latent space of pre-trained StyleGANs for image editing. Additionally, other works \cite{pivotal} go a step further and fine-tune the pre-trained StyleGANs on the given image to better preserve visual details during editing.
To achieve text-driven image editing, some researchers \cite{paint,stylegan,clip,clip2stylegan,gal2022stylegan} utilize pre-trained GAN generators \cite{stylegan2} and text encoders \cite{clip} to progressively optimize an image based on a given text prompt. Despite the encouraging outcomes, GAN-based methods are often constrained to edit images from a specific domain for which the GAN was initially trained.  Unlike GAN-based methods, we aim to harness the powerful generative capabilities of existing large-scale text-to-image diffusion models to develop an image editing method suitable for various domains.

\subsection{DDPM for image editing}
Over the past few years, diffusion models have developed rapidly since the advent of denoising diffusion probabilistic models (DDPM) \cite{ddpm}, achieving state-of-the-art performance in terms of image quality and mode coverage. Large-scale diffusion models, such as Imagen \cite{imagen}, DALL$\cdot$E3 \cite{dalle3},  GEN-1 \cite{gen-1}, Emu Video \cite{emuvideo}, and Stable Diffusion \cite{ldm}, have greatly enhanced the ability to generate images or videos from plain text. Naturally, \redd{recent studies \cite{dreambooth,imagic,bldm,bdm,diffedit,sdedit,cds,huang2024diffusion,huang2024sbcr,huang2024entwined,chen2023specref,huang2024wavedm,lv2024gpt4motion}} have focused on exploiting the rich semantic knowledge embedded in text-guided diffusion models for realistic image editing.

A slew of studies \cite{dreambooth,imagic,textual-inversion,diffusionclip} propose learning a unique word for a given object and fine-tuning the diffusion models case-specifically for different text prompts. These methods usually introduce a case-specific optimization process, which may limit their applications in real-time.  \redd{PowerPaint \cite{ppt}, however, improves this approach by introducing learnable task prompts and tailored fine-tuning to direct the model's focus across different image inpainting tasks, thereby reducing the need for concept-specific model adjustments but still requiring model fine-tuning.}

In contrast, another liner of research focuses on designing \redd{optimization-free methods \cite{bdm,ilvr,sdedit,bldm,diffedit,repaint,huang2024entwined,ledits}}, which require no fine-tuning or optimization and can be easily extended to any pre-trained diffusion models. To preserve the details of original images, these methods typically perform spatial blending between the noised target images and the text-guided diffusion latent variables at each time step of the denoising process, using either a user-provided mask \cite{bdm,bldm,sdedit}  or a self-predicted mask \cite{diffedit}.

\redd{As an optimization-free approach, PFB-Diff stands out  by manipulating deep feature maps within noise prediction networks, rather than simply editing intermediate noisy images.
	Deep features contain rich semantics \cite{hittawe2019abnormal, afzal2023visualization, hittawe2022efficient, HARROU2022197}, allowing feature-level editing to effectively preserve semantic consistency in edited images. Furthermore, this paper is the first to identify and address the challenge of mutual interference in multi-object editing scenarios, where previous methods typically fail to generate multiple distinct objects simultaneously.}

\subsection{Attention control for image editing}
Prompt-to-Prompt \cite{ptp} controls the editing of synthesized images by manipulating the cross-attention maps; however, its editing ability is limited when applied to real images. When combined with Null-text inversion \cite{null-text}, i.e., an accurate inversion technique, Prompt-to-Prompt \cite{ptp} can be extended to real image editing. Pix2pix-zero \cite{pix2pix} goes one step further than Prompt-to-Prompt \cite{ptp} by optimizing the cross-attention maps during denoising to approximate the given reference cross-attention maps, thereby achieving better content preservation and image quality. \redd{Recently, SpecRef \cite{chen2023specref} introduces a specialized reference attention controller designed to integrate features from a reference image into target images. While attention editing techniques have been employed in previous works, they predominantly focus on leveraging attention maps for image translation or mask prediction, without exploring the role of attention in multi-object replacement scenarios. PFB-Diff is the first to restrict specific words' attention maps to prevent interference between objects during multi-object generation, allowing for accurate object placement at user-specified locations.}

Although previous works employ attention editing techniques, they mainly focus on leveraging attention maps for image translation or mask prediction, without addressing multi-object replacement scenarios. PFB-Diff is the first to restrict specific words' attention maps to prevent interference between objects during multi-object generation, enabling accurate object generation at user-specified locations.

\section{Preliminaries} \label{sec:background}

\paragraph{Diffusion models} Denoising diffusion probabilistic models (DDPMs) \cite{ddpm} is a class of generative models. Given a set of real data $\mathbf{x}_0\sim q(\mathbf{x}_0)$, diffusion models aim to approximate the data distribution $q(\mathbf{x}_0)$ and sample from it. The diffusion model consists of a forward diffusion process that gradually injects noise into the data and a backward denoising process that aims to generate data from the noise. The forward noise-injection process is formalized as a Markov chain  with Gaussian transitions:
\begin{align}
	\begin{split}
		q\left(\mathbf{x}_{1: T}\right|\mathbf{x}_{0})&=q\left(\mathbf{x}_{0}\right) \prod_{t=1}^{T} q\left(\mathbf{x}_{t} | \mathbf{x}_{t-1}\right),\\
		q\left(\mathbf{x}_{t}|\mathbf{x}_{t-1}\right)&=\mathcal{N}\left(\mathbf{x}_{t} ; \sqrt{1-\beta_{t}} \mathbf{x}_{t-1}, \beta_{t} \mathbf{I}\right), 
	\end{split}
\end{align}
where $\beta_t \in(0,1)$ represents the noise schedule at time $t$. When \(T\) is sufficiently large, the latent variable $\mathbf{x}_T$ approximates an isotropic Gaussian distribution. The objective of the reverse denoising process is to reconstruct the true sample from a Gaussian noise input, $\mathbf{x}_{T} \sim \mathcal{N}(\mathbf{0}, \mathbf{I})$, by sampling from $q(\mathbf{x}_{t-1} | \mathbf{x}_{t})$ sequentially.
Note that if $\beta_{t}$ is sufficiently small, the conditional distribution $q(\mathbf{x}_{t-1} | \mathbf{x}_{t})$ will also be a Gaussian distribution.
Unfortunately, estimating $q(\mathbf{x}_{t-1} | \mathbf{x}_{t})$  is challenging due to its dependence on the intractable data distribution $q(\mathbf{x}_0)$. 
Hence, the conditional distribution $q(\mathbf{x}_{t-1} | \mathbf{x}_{t})$ is approximated by a Gaussian distribution whose mean and covariance are predicted by a deep neural network $p_{\theta}$. Then, $\mathbf{x}_{t-1}$ is sampled from the Gaussian distribution based on the predicted parameters,
\begin{equation}
	p_{\theta}\left(\mathbf{x}_{t-1}| \mathbf{x}_{t}\right)=\mathcal{N}\left(\mathbf{x}_{t-1} ; \boldsymbol{\mu}_{\theta}\left(\mathbf{x}_{t}, t\right), \boldsymbol{\Sigma}_{\theta}\left(\mathbf{x}_{t}, t\right)\right).
	\label{eq:reverse}
\end{equation}
Instead of directly estimating $\boldsymbol{\mu}_{\theta}\left(\mathbf{x}_{t}, t\right)$, Ho et al. \cite{ddpm} suggest predicting the noise  $\epsilon_\theta \left(\mathbf{x}_t, t \right)$ that was introduced to $\mathbf{x}_0$ to produce $\mathbf{x}_t$, following the objective:
\begin{equation}
	\min _\theta E_{\mathbf{x}_0, \boldsymbol{\varepsilon} \sim\mathcal{N}(\mathbf{0}, \mathbf{I}), t \sim \operatorname{Uniform}(1, T)}\left\|\boldsymbol{\varepsilon}-\epsilon_\theta \left(\mathbf{x}_t, t \right)\right\|_2^2.
\end{equation}
Subsequently, $\boldsymbol{\mu}_{\theta}\left(\mathbf{x}_{t}\right)$ can be derived using Bayes' theorem,
\begin{equation}
	\boldsymbol{\mu}_{\theta}\left(\mathbf{x}_{t}, t\right)=\frac{1}{\sqrt{\alpha_t}}\left( \mathbf{x}_t - \frac{\beta_t}{\sqrt{1-\bar{\alpha}_t}}\boldsymbol{\epsilon}_\theta(\mathbf{x}_t,t) \right),
\end{equation}
where $\alpha_t=1-\beta_t$ and $\bar{\alpha}_t=\prod_{i=1}^t\alpha_i$.
At inference time, we start the backward process from a random noise \( \mathbf{x}_{T} \sim \mathcal{N}(\mathbf{0}, \mathbf{I}) \) and iteratively apply Eq.~(\ref{eq:reverse}) to derive \( \mathbf{x}_{t-1} \) from \( \mathbf{x}_{t} \) until \( t = 0 \). For more details of DDPMs, please refer to \cite{ddpm,improved-ddpm,sohl2015deep}.
\paragraph{DDIM sampling}
Text-guided diffusion models are designed to transform a random noise $\mathbf{x}_{T}$ and a text condition $\mathcal{C}$ into an output image $\mathbf{x}_{0}$ aligned with the given text prompt. To facilitate faster sampling in the reverse process, 
we utilize the deterministic DDIM \cite{ddim} sampling to sequentially remove the noise:
\begin{equation}
	\mathbf{x}_{t-1}=\sqrt{\frac{\alpha_{t-1}}{\alpha_t}} \mathbf{x}_{t}+\left(\sqrt{\frac{1}{\alpha_{t-1}}-1}-\sqrt{\frac{1}{\alpha_t}-1}\right) \epsilon_\theta\left(\mathbf{x}_{t}, t, \mathcal{C}\right).
	\label{eq:ddim-sampling}
\end{equation}
Eq.~(\ref{eq:ddim-sampling}) can be written as the neural ODE, taking $\mathbf{u}=\mathbf{x} / \sqrt{\alpha}$ and $\tau=\sqrt{1 / \alpha-1}$ :
\begin{equation}
	d \mathbf{u}=\epsilon_\theta\left(\frac{\mathbf{u}}{\sqrt{1+\tau^2}}, t\right) d \tau.
	\label{eq:ode}
\end{equation}
This allows us to view DDIM sampling as an Euler scheme for solving Eq.~(\ref{eq:ode}) with initial condition $\mathbf{u}(t=T) \sim \mathcal{N}\left(\mathbf{0}, \alpha_T \mathbf{I}\right)$. 
As proposed by Song et al. \cite{ddim}, we can also use this ODE to encode an image $\mathbf{x}_{0}$ onto a latent variable $\mathbf{x}_{r}$ for a timestep $r$,  based on the assumption that the ODE process is reversible in infinitesimal steps:
\begin{equation}
	\mathbf{x}_{t+1}=\sqrt{\frac{\alpha_{t+1}}{\alpha_t}} \mathbf{x}_{t}+\left(\sqrt{\frac{1}{\alpha_{t+1}}-1}-\sqrt{\frac{1}{\alpha_t}-1}\right) \epsilon_\theta\left(\mathbf{x}_{t}, t, \mathcal{C}\right).
	\label{eq:normal-encode} 
\end{equation}
In other words, the diffusion process is performed in the reverse direction, that is, $\mathbf{x}_{0} \rightarrow \mathbf{x}_{T}$ instead of $\mathbf{x}_{T} \rightarrow \mathbf{x}_{0}$.
\paragraph{Latent diffusion models}
Diffusion models often operate in the RGB space where $\mathbf{x}_0$  is a pixel image. In this paper, we build our approach on the popular latent diffusion model (LDM)  \cite{ldm}. Specifically, the input image is first mapped to the latent encoding $\mathbf{x}_0$ by an image encoder. Then, the forward noise-injection process and the backward denoising process of the diffusion model are performed in this latent space. At the end of the backward denoising process, an image decoder is employed to restore $\mathbf{x}_0$ back to the image.
For simplicity, we refer to the latent image encoding $\mathbf{x}_0$ as ``image" in the following sections. To build more flexible conditional image generators, LDM augments the underlying U-Net \cite{unet} backbone with the cross-attention \cite{attn} mechanism. The noise prediction network $\epsilon_{\theta}$ in LDM is a U-shaped network composed of residual and transformer blocks. 
\begin{figure*}[t]
	\centering
	\includegraphics[width=\textwidth]{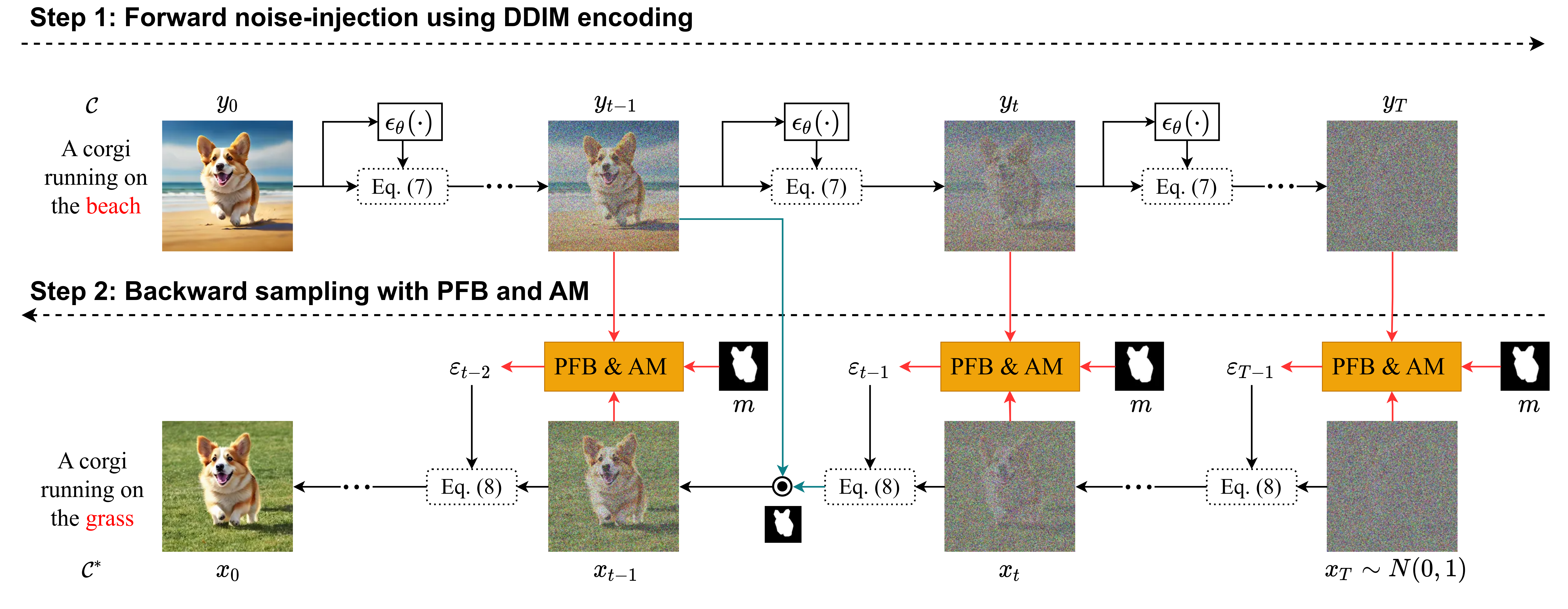}
	\caption{Pipeline of the proposed PFB-Diff framework. Our method utilizes a pre-trained diffusion model without additional training. Initially, we employ DDIM encoding to obtain noisy images $\mathbf{y}_t$ for all time steps. Subsequently, we perform a backward denoising process starting from a Gaussian noise $\mathbf{x}_T$ to generate the edited image $\mathbf{x}_0$. When estimating the noise $\boldsymbol{\varepsilon}_t$ at each time step, we adopt progressive feature blending (indicated by the red lines) to fuse the deep features of $\mathbf{x}_t$ and $\mathbf{y}_t$. Note that we also incorporate intermediate noisy image blending (indicated by blue lines) in the early stages of the denoising process to better preserve visual details in irrelevant regions. We denote with $\odot$ the element-wise blending of two images using the input mask $\mathbf{m}$. The Eq.~(7) and Eq.~(8) in the figure refer to Eq.~(\ref{eq:normal-encode}) in Section \ref{sec:background} and Eq.~(\ref{eq:overview}) in Section \ref{sec:me-overview}, respectively. Better viewed online in color and zoomed in for details.}
	\label{fig:overview}
\end{figure*}
\section{Methods}
In this section, we first give an overview of PFB-Diff (Section \ref{sec:me-overview}). We then describe two key components of our method, i.e., progressive feature blending (Section \ref{sec:me-pfb}) and attention masking mechanism (Section \ref{sec:me-ma}), in detail.
\subsection{An overview of our PFB-Diff framework} \label{sec:me-overview}
Given an input image $\mathbf{y}_0$, along with its corresponding coarse text description $d$, a target text prompt $d^*$, and a binary mask $\mathbf{m}$ indicating the region of interest in the image, our objective is to generate a manipulated image $\mathbf{x}_0$ that fulfills the following requirements. Firstly, the entire image should be aligned with the target text prompt. Secondly, the uninterested regions should remain unaltered. Thirdly, the edited result $\mathbf{x}_0$ should demonstrate high consistency and quality.

To fulfill the above requirements, we propose a Progressive Feature Blending method for Diffusion-based image editing, dubbed PFB-Diff. An overview of our method is given in Figure \ref{fig:overview}.
Formally, given an input image-text pair ($\mathbf{y}_0$, $d$), a target text prompt $d^*$, and a binary mask $\mathbf{m}$ indicating the regions of interest, we first encode the text prompts $d$ and $d^*$ into embeddings $\mathcal{C}$ and $\mathcal{C}^*$ respectively, using a pre-trained text encoder. Next, we sample the noisy image $\mathbf{y}_t$ at each time step with DDIM encoding 
using the pre-trained diffusion model $\epsilon_{\theta}$. Finally, starting from a random Gaussian noise $\mathbf{x}_T$, we iteratively sample $\mathbf{x}_t$ using the following equation until $t=0$:
\begin{align}
	\mathbf{x}_{t-1}=\sqrt{\frac{\alpha_{t-1}}{\alpha_t}} \mathbf{x}_{t}+ \left(\sqrt{\frac{1}{\alpha_{t-1}}-1}-\sqrt{\frac{1}{\alpha_t}-1}\right) \cdot \boldsymbol{\varepsilon}_{t-1},
	\label{eq:overview}
\end{align}
where $\boldsymbol{\varepsilon}_{t-1}$ represents the estimated noise at time $t-1$. A key feature of PFB-Diff is that the text-driven blended diffusion is made not directly on the intermediate noisy images but instead on their correspondingly learned feature maps. To accomplish this, we introduce a Progressive Feature Blending (PFB) technique. Additionally, we introduce an Attention Masking (AM) mechanism for cross-attention layers, enabling text-guided generation within specific regions of interest. By incorporating both the PFB and AM into the noise prediction network $\epsilon_{\theta}$, we obtain the diffusion model $\hat{\epsilon}_{\theta}$. The $\boldsymbol{\varepsilon}_{t-1}$ in Eq.~(\ref{eq:overview}) is derived from $\hat{\epsilon}_{\theta}$ as follows,
\begin{equation}
	\boldsymbol{\varepsilon}_{t-1} = \hat{\epsilon}_\theta\left(\mathbf{x}_t, t, \mathcal{C}^*,{\mathbf{y}}_{t}, \mathcal{C},\mathbf{m}\right).
\end{equation}
To further preserve the details of irrelevant regions, we also apply latent-level blending during the early stages of the denoising process, illustrated by the blue lines in Figure \ref{fig:overview}.

\subsection{Progressive Feature Blending (PFB)} \label{sec:me-pfb}

\begin{figure}[t]
	\centering
	\includegraphics[width=\columnwidth]{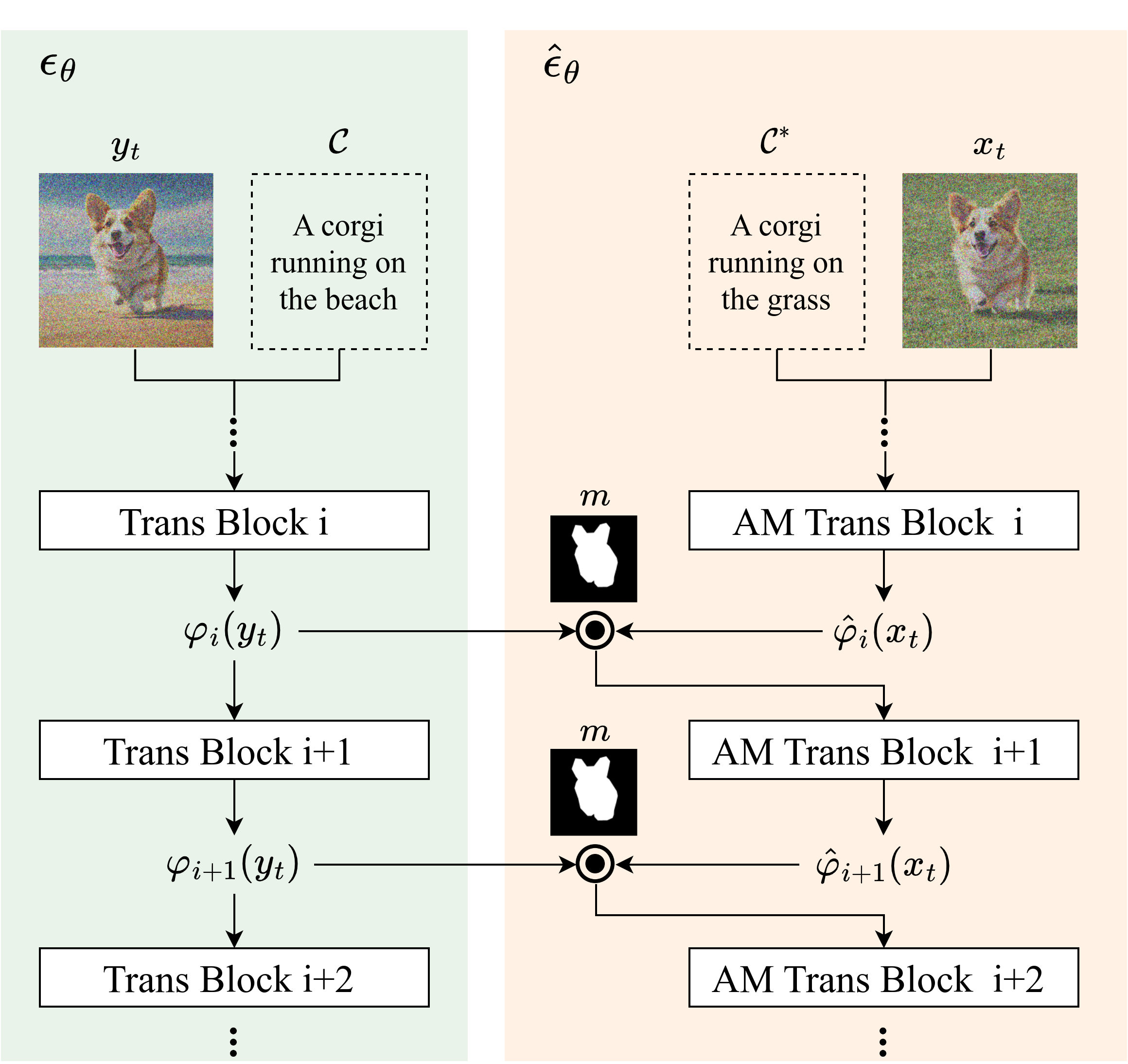}
	\caption{Progressive feature blending. The left shows the pre-trained noise prediction network $\epsilon_{\theta}(\cdot)$ in LDM, which is built upon U-Net \cite{unet} architecture and composed of transformer blocks. The right $\hat{\epsilon}_{\theta}$ is obtained by plugging progressive feature blending (PFB) and attention masking (AM) into $\epsilon_{\theta}$. ``Trans Block i" indicates the i-th transformer block in $\epsilon_{\theta}(\cdot)$. ``AM Trans Block i" indicates the i-th transformer block (equipped with attention masking) in $\hat{\epsilon}_{\theta}(\cdot)$. For more details of the attention masking mechanism, please refer to Section \ref{sec:me-ma}. Note that the left $\epsilon_{\theta}$ and the right $\hat{\epsilon}_{\theta}$ share the same weights.  We denote with $\odot$ the element-wise blending of two feature maps using the provided mask $\mathbf{m}$.}
	\label{fig:pfb}
\end{figure}

To achieve localized editing, previous methods \cite{bdm, bldm, sdedit, repaint,diffedit} typically involve blending the generated intermediate latent variables with noised versions of the input image at different noise levels. However, empirical experiments indicate that this approach can introduce artifacts and inconsistencies in the results. This is because the low-level information in the intermediate noisy images lacks the necessary semantics to produce consistent and seamless fusion. While extensions have been made in latent diffusion \cite{ldm}, which leverages a pre-trained autoencoder to encode the input image into a lower-dimensional hidden space for efficient diffusion, the hidden space itself still lacks the essential semantics required to maintain image consistency. To address these limitations, we propose a simple yet effective module called Progressive Feature Blending (PFB). This module can be plugged into any pre-trained text-to-image frameworks to enhance their ability to generate consistent and high-quality images.

Recall that each diffusion step $t$ contains predicting the noise $\boldsymbol{\varepsilon}$ with ($x_t$, $\mathcal{C}^*$, $\mathbf{y}_t$, $\mathcal{C}$, $\mathbf{m}$) using a U-shaped network $\hat{\epsilon}_\theta$ composed of transformer blocks. In the left branch of Figure \ref{fig:pfb}, we show three consecutive blocks of the pre-trained noise prediction network $\epsilon_{\theta}$ for $\mathbf{y}_t$ and denote the $i$-th transformer block's output as $\varphi_i(\mathbf{y}_t)$. PFB is implemented by inserting the feature blending module into certain layers of the noise prediction network, as shown in the right branch of Figure \ref{fig:pfb}.
At each layer, we blend the feature map $\hat{\varphi}_i(\mathbf{x}_t)$ with $\varphi_i(\mathbf{y}_t)$ using the binary mask $\mathbf{m}$ as below:
\begin{equation}
	\widetilde{\varphi}_i(\mathbf{x}_t) = \varphi_i(\mathbf{y}_t)  \odot (1-\mathbf{m}) + \hat{\varphi}_i(\mathbf{x}_t) \odot \mathbf{m},
	\label{eq:feature-blend}
\end{equation}
where blended feature map $\widetilde{\varphi}_i(\mathbf{x}_t)$, rather than $\hat{\varphi}_i(\mathbf{x}_t)$, is fed into the next transformer block. Since the feature map size changes from block to block, we downsample or upsample the binary mask $\mathbf{m}$ accordingly to make the computation of Eq.~(\ref{eq:feature-blend}) valid.
Note that we also modify the transformer block by incorporating an Attention Masking (AM) mechanism into the cross-attention layers, with details described in Section \ref{sec:me-ma}.

To summarize, the PFB module implements a new, feature-level text-driven blended diffusion, leading to high-quality, seamless target generation towards the guiding text prompt.
\begin{figure}[t]
	\centering
	\includegraphics[width=\columnwidth]{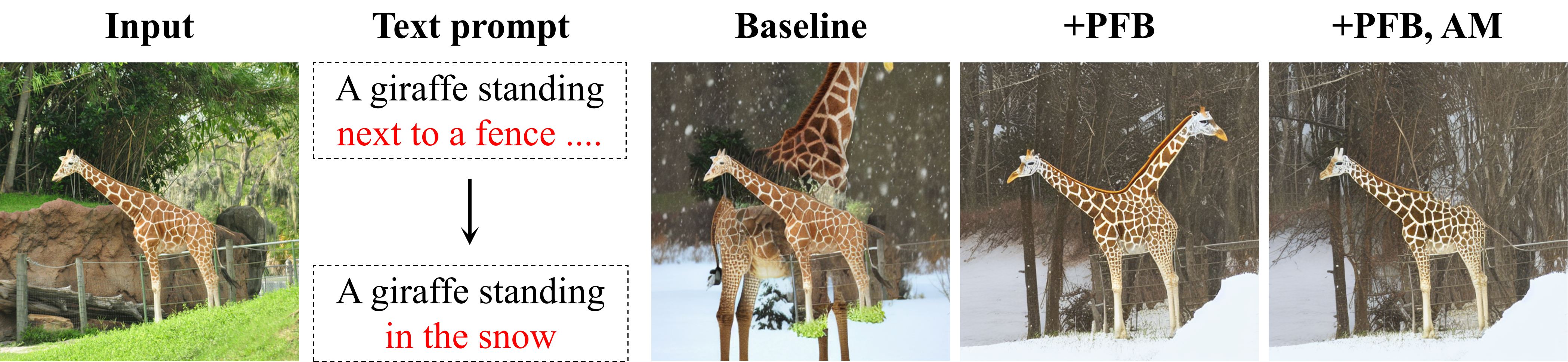}
	\caption{The role of Attention Masking (AM) in background replacement. The first column showcases a sample from the COCO dataset \cite{coco}. The second column presents the target text prompt. In the third column, without PFB and AM, previous latent-level blending methods introduce an additional giraffe in the background. In the fourth column, by utilizing PFB, the model generates the giraffe to the desired location, but with a clear difference from the foreground instance. In the fifth column, AM ensures that the generated giraffe is faithful to the original one.}
	\label{fig:attn}
\end{figure}

\begin{figure}[t]
	\centering
	\includegraphics[width=\columnwidth]{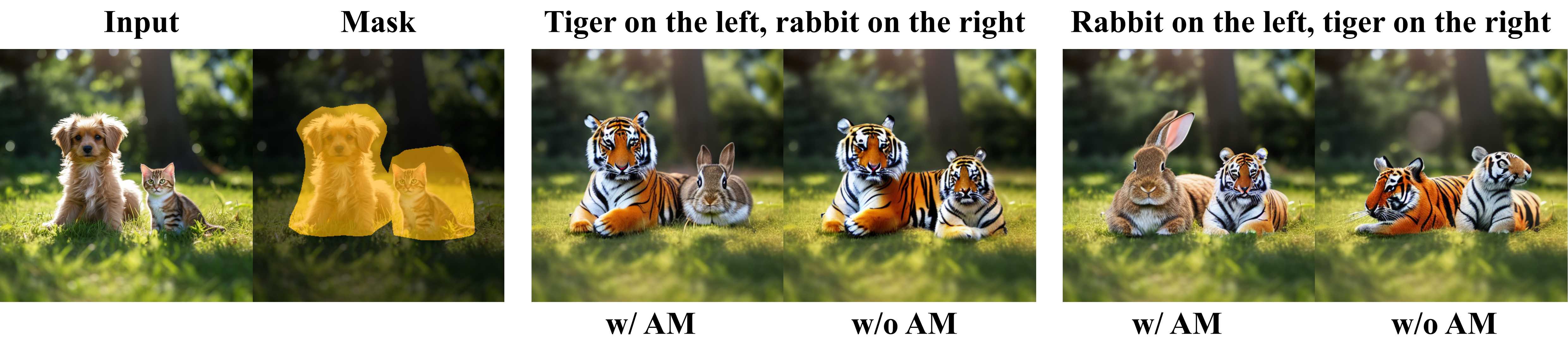}
	\caption{\red{The role of Attention Masking (AM) in multi-object replacement. ``w/o AM" and ``w/ AM" represent the image editing results without and with the use of the Attention Masking (AM) mechanism, respectively.}}
	\label{pfig:me-am-2}
\end{figure}

\begin{figure*}[t]
	\centering
	\includegraphics[width=0.85\textwidth]{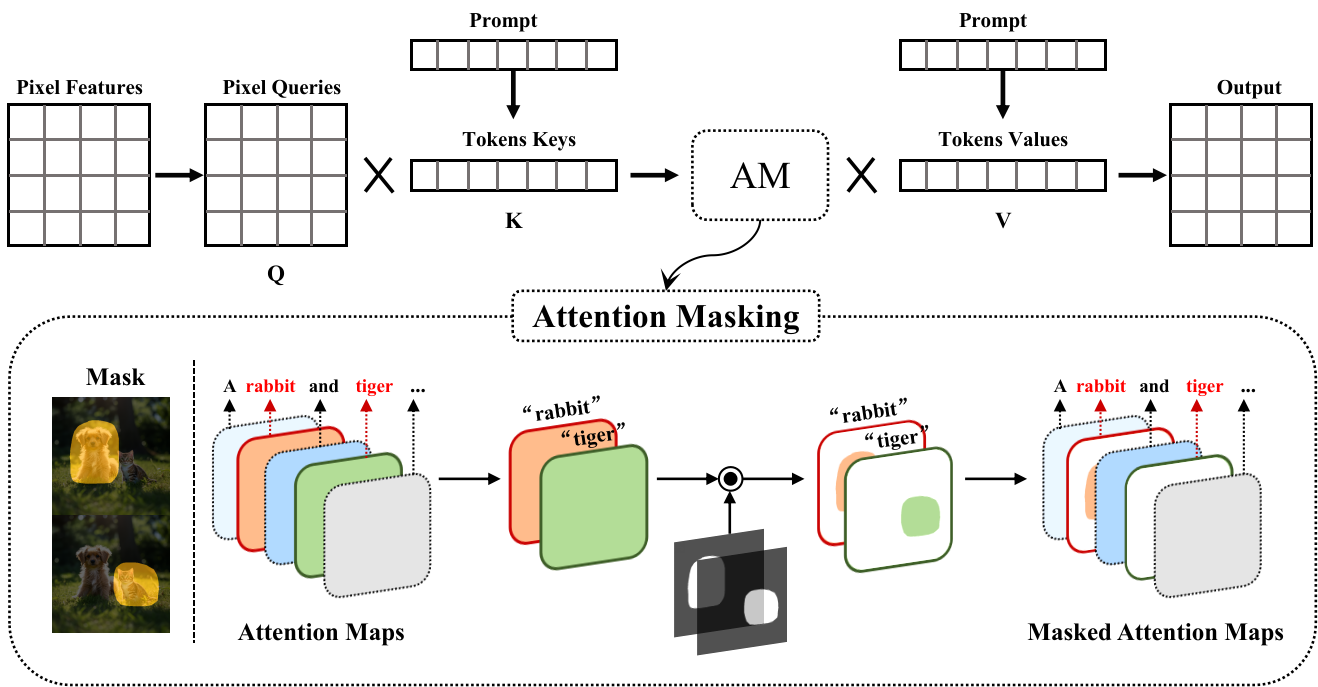}
	\caption{\red{The details of the Attention Masking (AM) mechanism in the cross-attention layers. Each channel of the attention map is associated with a specific word. The channels corresponding to ``rabbit" and ``tiger" are modified using the input mask to restrict their influence on image features to specified areas. In the figure, $\odot$ represents element-wise multiplication.}}
	\label{fig:ma}
\end{figure*}

\begin{figure*}[t]
	\centering
	\includegraphics[width=0.85\textwidth]{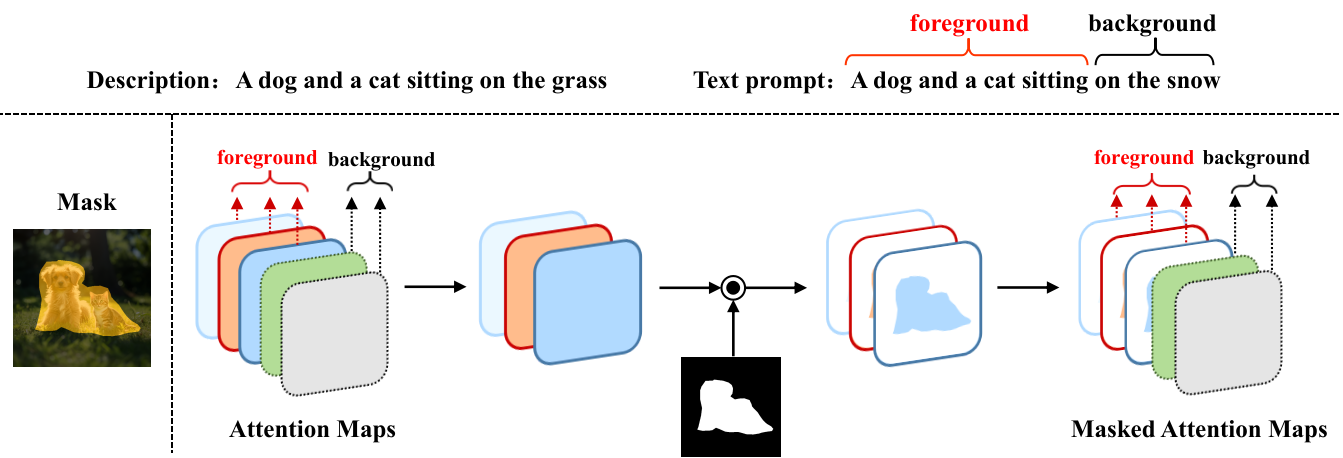}
	\caption{\red{Attention masking mechanism in background replacement}}
	\label{pfig:me-amm-1}
\end{figure*}
\subsection{Attention Masking (AM) mechanism} \label{sec:me-ma}
When manipulating an input image from ``a giraffe stands by the fence" to ``a giraffe stands in the snow", choosing ``snow" as the target prompt may lead to a generated image lacking global semantic consistency. Instead, we input the global text prompt ``a giraffe stands in the snow" into the diffusion model to generate a new background. However, latent-level blending cannot effectively constrain the generated giraffe to its original location. As shown in the third column (baseline) in Figure \ref{fig:attn}, blending the noised input image with newly generated content using the input binary mask can result in noticeable artifacts. While the proposed progressive feature blending helps to restrict the generated giraffe to desired regions, as shown in the fourth column of Figure \ref{fig:attn}, the model still sometimes produces unwanted artifacts around the foreground giraffe.

\red{In multi-object replacement scenarios, users typically specify a desired generation location for each target object. However, due to the limited capabilities of Stable Diffusion \cite{ldm} in language understanding and image generation, the model often fails to accurately generate target objects at the specified locations based solely on the location information in the text prompt. Additionally, as shown in columns 4 and 6 of Figure ~\ref{pfig:me-am-2}, when multiple objects need to be generated simultaneously, their mutual interference often results in some objects not being generated correctly.}

Masking the attention maps of specific words is introduced to address these issues.
As mentioned in Figure~\ref{fig:pfb}, we present an Attention Masking (AM) mechanism to fuse the visual features with the textual ones. The details of the attention masking mechanism are provided in Figure~\ref{fig:ma}. Specifically, the deep spatial features of the noisy image are projected to a query matrix $Q$, and the text prompt's embedding is projected to a key matrix $K$ and a value matrix $V$ via learned linear projections. The attention score map is computed by  
\begin{equation}
	M = \left(QK^T/\sqrt{n} \right.),
\end{equation}
where $n$ is the latent projection dimension.

\red{In scenarios involving the replacement of multiple objects, the users need to specify the desired location of each target object by providing ``object-mask" pairs. These pairs define the generation area for each target object. As illustrated in Figure~\ref{fig:ma}, the Attention Masking (AM) module modifies the attention map for each target object (word) based on the corresponding mask. Given the binary mask $\mathbf{m}$, the attention masking mechanism adjusts the attention score matrix as follows: }
\begin{equation}
	\widehat{M^k_{ij}} = \left\{
	\begin{aligned}
		M^k_{ij} & , & \mathbf{m}^k_{ij}=1, \\
		-inf & , & \mathbf{m}^k_{ij}=0,
	\end{aligned}
	\right.
\end{equation}
where the entry $M^k_{ij}$ of the matrix $M$ gives the weight of the value of the $k$-th textual token at the position $(i,j)$. Finally, the output of the masked cross-attention layer is defined as 
\begin{equation}
	\mathcal{F} = \text{softmax}(\widehat{M})V.
\end{equation}

By utilizing masked attention maps, the model can effectively limit the impact of each word on the image features to specific spatial regions. As a result, the model can enforce the generation of the target objects to desired positions and shapes.

\red{As shown in Figure~\ref{pfig:me-amm-1}, in background replacement, the user does not need to provide a separate mask for each foreground object, but only needs to provide a single mask that labels the entire foreground.}

\subsection{\redd{Implementation}}   \label{sec:our-detail}

\color{black}
\begin{algorithm}[t]
	\caption{\redd{PFB-Diff for Image Editing}}
	\label{alg:pfbdiff}
	\color{black}
	\begin{algorithmic}[1]
		\Require Input image $\mathbf{y}_{0}$, target edit condition $\mathcal{C}^*$, original description $\mathcal{C}$, mask $\mathbf{m}$ indicating area of interest, pre-trained diffusion model $\epsilon_\theta$, enhanced model $\hat{\epsilon}_\theta$ with Progressive Feature Blending (PFB) and Attention Masking (AM) techniques, number of diffusion steps $T$,  termination timestep for pixel-level blending $r$, and PFB \& AM deactivation timestep $s$
		\Ensure Edited image $\mathbf{x}_{0}$ 
		
		\State // Step 1: Encode the input image into a sequence of noisy images
		\For{$t = 0$ to $T$}
		\State Predict noise $\boldsymbol{\varepsilon}_{t}$ from the current noisy image $\mathbf{y}_{t}$ using $\epsilon_\theta(\mathbf{y}_{t}, t, \mathcal{C})$
		\State Update $\mathbf{y}_{t+1}$ with the predicted noise: $\mathbf{y}_{t+1}=\sqrt{\frac{\alpha_{t+1}}{\alpha_t}} \mathbf{y}_{t} + \left(\sqrt{\frac{1}{\alpha_{t+1}} - 1} - \sqrt{\frac{1}{\alpha_t} - 1}\right) \boldsymbol{\varepsilon}_{t}$
		\EndFor
		
		\State // Step 2: Reverse the process to generate the edited image
		\State Initialize the denoising process with random noise $\mathbf{x}_T \sim \mathcal{N}(0, I)$
		\For{$t = T$ to $1$}
		\If{$t > s$}
		\State Predict noise $\boldsymbol{\varepsilon}_{t-1}$ using $\hat{\epsilon}_\theta(\mathbf{x}_t, t, \mathcal{C}^*, \mathbf{y}_t, \mathbf{m})$ with  PFB and AM
		\State Update $\mathbf{x}_{t-1}$ with the predicted noise: $\mathbf{x}_{t-1} = \sqrt{\frac{\alpha_{t-1}}{\alpha_t}} \mathbf{x}_{t} + \left(\sqrt{\frac{1}{\alpha_{t-1}} - 1} - \sqrt{\frac{1}{\alpha_t} - 1}\right) \boldsymbol{\varepsilon}_{t-1}$
		\Else
		\State Predict noise $\boldsymbol{\varepsilon}_{t-1}$ using $\epsilon_\theta(\mathbf{x}_t, t, \mathcal{C}^*)$
		\State Update $\mathbf{x}_{t-1}$ similarly as above
		\EndIf
		\If{$t > r$}
		\State Blend the original and new content using the mask: $\mathbf{x}_{t-1} = \mathbf{m} \mathbf{x}_{t-1} + (1 - \mathbf{m}) \mathbf{y}_{t-1}$ 
		\EndIf
		\EndFor
		\State \Return the final edited image $\mathbf{x}_{0}$ 
	\end{algorithmic}
\end{algorithm}

\color{black}

\begin{color}{black}
	Given an input real image, we first utilize DDIM encoding to obtain intermediate noisy images, represented as $\mathbf{y}_t$ in Figure \ref{fig:overview}. Afterward, we perform the reverse denoising process from a randomly generated Gaussian noise $\mathbf{x}_T$ and progressively denoise it. The detailed implementation of PFB-Diff is outlined in Algorithm \ref{alg:pfbdiff}.
	
	When estimating noise at each time step, we incorporate progressive feature blending (PFB) and attention masking (AM) in specific layers of the noise prediction model $\epsilon_\theta(\cdot)$. The model follows a U-Net \cite{unet} architecture consisting of 13 layers, each comprising a residual block \cite{resnet} and a transformer block \cite{attn}. Progressive feature blending is applied from layers 8 to 13 while attention masking is employed from layers 4 to 13.
	
	Furthermore, to enhance the faithfulness to the original images, we combine our approach with pixel-level blending during the early stages of the denoising process. This involves blending the noisy image $\mathbf{y}_t$ with the generated latent variable $\mathbf{x}_t$.  For object replacement, we apply pixel-level blending during the first 50\% of the timesteps, while for background replacement, it is applied for the first 20\% timesteps.

	Due to the inherent inaccuracies in user-provided masks, it's challenging to segment foreground objects without inadvertently incorporating elements of the background. For example, when isolating a horse from its grassy surroundings, parts of the grass are often included in the extracted mask. To mitigate this issue during background replacement, we employ a strategy of discontinuing progressive feature blending (PFB) and attention masking (AM) in the final 20\% of the timesteps. This early termination approach allows the diffusion model to finely tune the image, seamlessly bridging any gaps between the foreground object and the newly generated background, resulting in a more natural integration.
\end{color}

\section{Experiments}

\subsection{Experimental setup} \label{sec:set-up}
\paragraph{Dataset}
We conduct experimental evaluations on two datasets. We first construct a dataset composed of synthetic images generated by Midjourney\footnote{\url{https://www.midjourney.com/home/?callbackUrl=\%2Fapp\%2F}}, a highly popular text-to-image model. We have designed more than 50 text prompts with relatively simple descriptions, such as ``a happy corgi running on the beach." These prompts take into account various factors, including the number of objects, image style, object attributes, secondary objects, and spatial relationships between objects, allowing us to evaluate the performance of editing methods in diverse scenarios. We conduct most qualitative experiments on this dataset in order to evaluate edits that involve changing the background, replacing objects, or changing object properties. To evaluate edits based on more complex text prompts and conduct quantitative comparisons, we collect 9,843 images from the COCO dataset \cite{coco} to build the COCO-animals-10k dataset, or COCOA-10k for short. The images in the dataset contain objects from 9 specific classes: dogs, cats, sheep, cows, horses, birds, elephants, zebras, and giraffes. By appropriately modifying the captions of the images, COCOA-10k can be used to evaluate foreground object replacement and background editing. Please refer to the Appendix for further details. 
\paragraph{Diffusion models}
In our experiments, we use the pre-trained latent diffusion models \cite{ldm} as the backbone. We evaluate the text-driven editing on 890M parameter text-conditional model trained on LAION-5B \cite{schuhmannlaion}, known as stable diffusion \cite{ldm}, at $512\times512$ resolution. We use 50 steps in DDIM sampling \cite{ddim} with a fixed schedule to generate images, which allows within 13 seconds on a single TITAN V GPU. Besides, we follow \cite{diffedit} and use DDIM encoding to obtain intermediate noisy images of the input image. 
\paragraph{Compared baselines}
We compare our method with \redd{eight} state-of-the-art text-driven editing methods. These include mask-based methods such as Blended Latent Diffusion (BLDM) \cite{bldm}, Stable-Diffusion-Inpainting \cite{ldm} (SD-Inpainting), ControlNet Inpainting \cite{controlnet} (ControlNet-Inpaint), \redd{PowerPaint \cite{ppt}} and mask-free methods like DiffEdit \cite{diffedit}, Prompt-to-Prompt (PTP) \cite{ptp}, \redd{LEDITS++ \cite{ledits} and Contrastive Denoising Score (CDS) \cite{cds}}. It's worth noting that PTP cannot be directly applied to edit real images. To overcome this limitation, we employ the Null-text inversion \cite{null-text} approach to invert the images, enabling the application of PTP for editing real images.  To ensure a fair comparison in the input setting, we enhance the mask-free baselines by incorporating user-provided masks. For PTP \cite{ptp}, DiffEdit \cite{diffedit}, \redd{LEDITS++ \cite{ledits} and CDS \cite{cds}}, we augment these methods with  user-provided masks for a fair comparison. The modified versions are referred to as PTP-mask, DiffEdit-mask, \redd{LEDITS++mask, and CDS-mask}.

\paragraph{Evaluation metrics}
Text-based semantic image editing requires meeting both requirements: 1) the generated image needs to align with the text prompts, and 2) the generated image needs to maintain high quality and realism.
\textcolor{black}{To measure these two aspects, we utilize the following four metrics to evaluate the generated images: 
	1) Accuracy. We use the pre-trained object detector, YOLOv6\footnote{\url{https://github.com/meituan/YOLOv6}} \cite{yolov6}, to detect whether the target object appears in the edited image, and use the confidence of the target object to measure the accuracy of the editing.
	2) CLIP Score \cite{clip}, which evaluates the alignment between guided text prompts and edited images. A higher CLIP score indicates better matching to the text descriptions. }
\begin{figure*}[t]
	\centering
	\includegraphics[width=\textwidth]{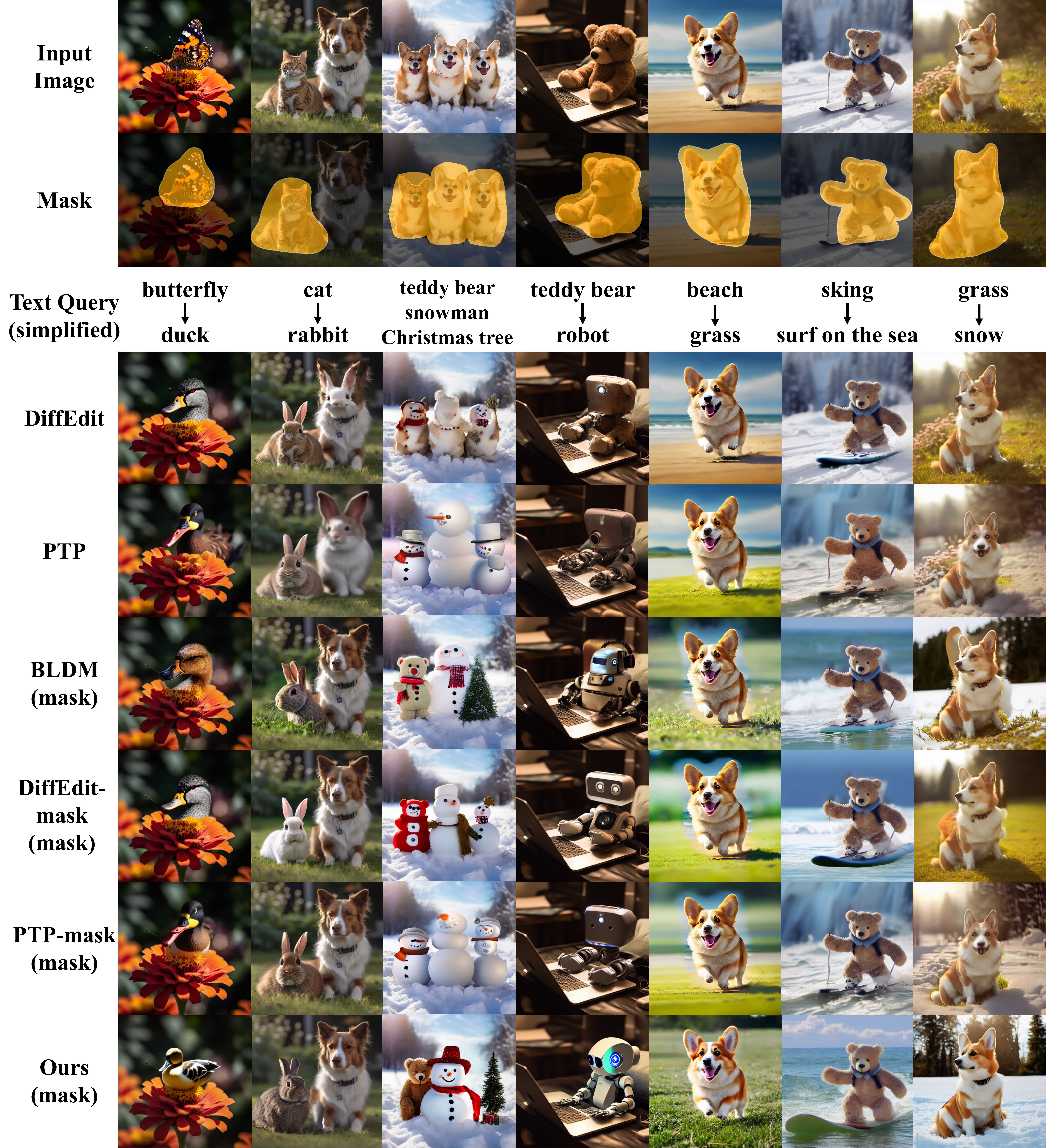}
	\caption{\red{Examples of edits on images obtained from Midjourney. For mask-based methods marked with ``(mask)", we use the manually annotated rough labels shown in the second row. For object/background replacement, DiffEdit \cite{diffedit} uses 100\% encoding rate during DDIM encoding. For PTP \cite{ptp}, we tried our best to adjust the hyperparameters for each image to obtain the best results. All the approaches adopt \texttt{Stable Diffusion v1-4} as the backbone, employing 50 steps of DDIM sampling. Better viewed online in color and zoomed in for details.} }
	\label{fig:exp-compare}
\end{figure*}
\textcolor{black}{3) Local CLIP Score, which evaluates the similarity between the local edit region and the target object. Since a single object occupies a small area in an image and corresponds to a single word in the text description, the global CLIP score cannot well reflect the quality of object replacement. To address this, we introduce Local CLIP Score, which concentrates on specific object regions by cropping images using bounding boxes. In the case of replacing a dog with a horse, we crop the target region from the edited image and evaluate its similarity to the text prompt ``a horse".
	4) CLIP-IQA \cite{clip-iqa}, which assesses both the quality perception (look) and abstract perception (feel) of images, leveraging rich visual language priors pre-encapsulated in CLIP \cite{clip} models.}

\subsection{Qualitative results}

\textcolor{black}{Figure \ref{fig:exp-compare} visualizes object and background editing on images generated by Midjourney\footnote{\url{https://www.midjourney.com/home/?callbackUrl=\%2Fapp\%2F}}. In comparison with other methods, PFB-Diff generally performs more targeted and accurate edits, leaving irrelevant regions intact and maintaining high image quality. By operating on deep features with rich semantics, our approach takes into account semantic consistency in the edited results. For instance, in the first column, only our method can seamlessly generate a complete duck; in the sixth column, our approach effectively replaces the snowboard with the surfboard, contrasting with other methods that tend to preserve some remnants of the original snowboard.} \red{As shown in the third column of Figure~\ref{fig:exp-compare}, in multi-object replacement scenarios, only BLDM~\cite{bldm} and PFB-Diff successfully generate multiple objects described in the text prompts, among which PFB-Diff exhibits superior visual effects. In contrast, other methods only manage to generate a snowman and the images suffer from significant artifacts.}

\begin{figure*}[t]
	\centering
	\includegraphics[width=0.9\textwidth]{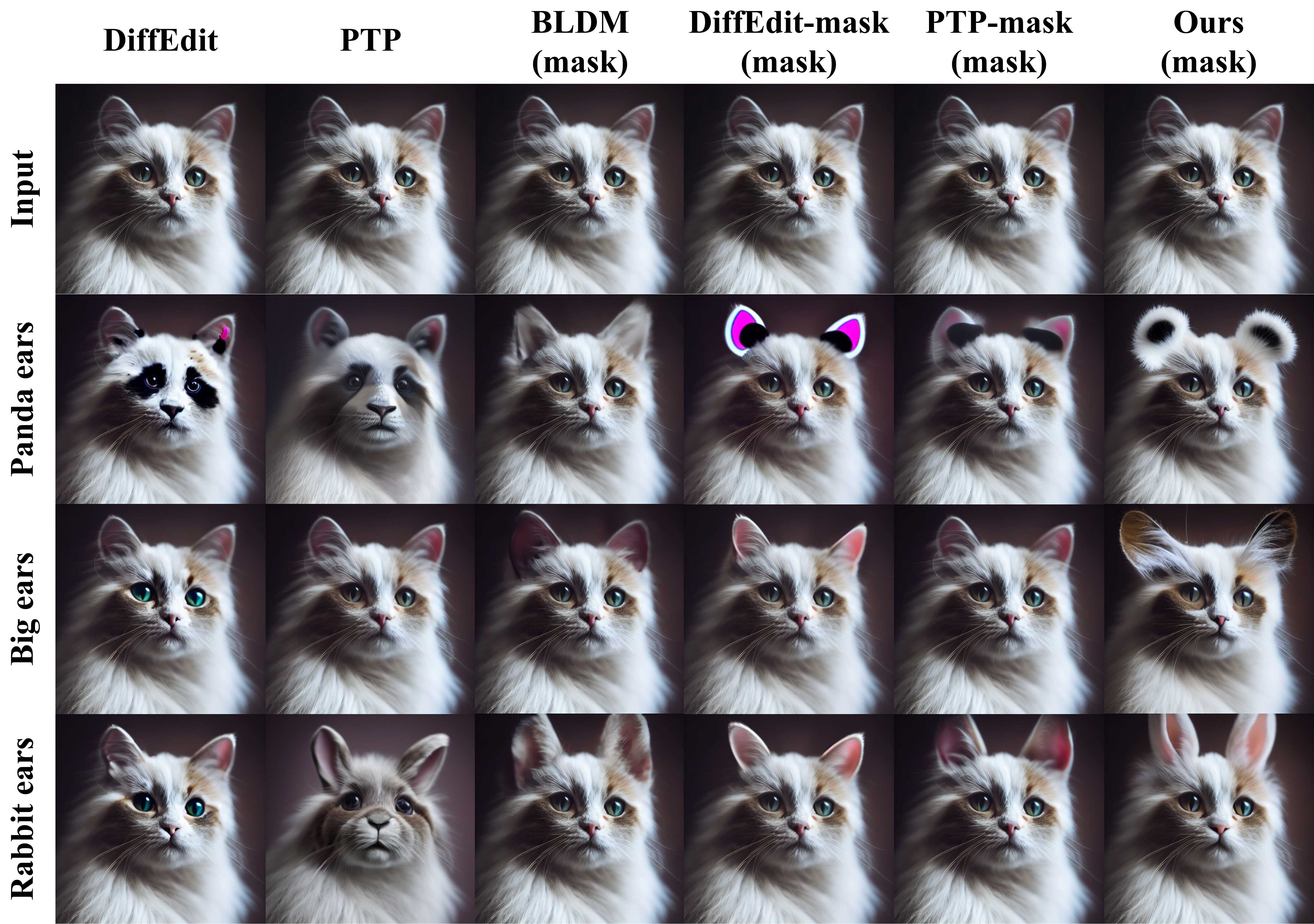}
	\caption{\textcolor{black}{Qualitative comparisons on cat ears editing. Our method can produce ears that match text prompts while others cannot. All the approaches adopt \texttt{Stable Diffusion v1-4} as the backbone, employing 50 steps of DDIM sampling.}}
	\label{fig:sup-cat}
\end{figure*}

DiffEdit \cite{diffedit} and PTP \cite{ptp} can cause undesired modifications and perform poorly when an image contains multiple objects, as shown in the second column of Figure \ref{fig:exp-compare}. Additionally, DiffEdit struggles to generate desired scenes in background replacement. After enhancing these methods with user-provided masks, the accuracy of object replacement has been improved, but the performance of background replacement and the perceptual quality of edited images remain unsatisfactory.  Although BLDM \cite{bldm} can accurately generate content aligned well with text prompts, it often lacks semantic consistency in edited results. 
Figure \ref{fig:sup-cat} further demonstrates the effectiveness of our method on editing only a part of an object, where only PFB-Diff can precisely modify the style of cat ears, while other methods cannot.  \redd{The qualitative results of CDS \cite{cds}, LEDITS++ \cite{ledits},  CDS-mask \cite{cds}, LEDITS++mask  \cite{ledits} and PowerPaint \cite{ppt} are detailed in the Appendix.}

\begin{figure*}[t]
	\centering
	\includegraphics[width=\textwidth]{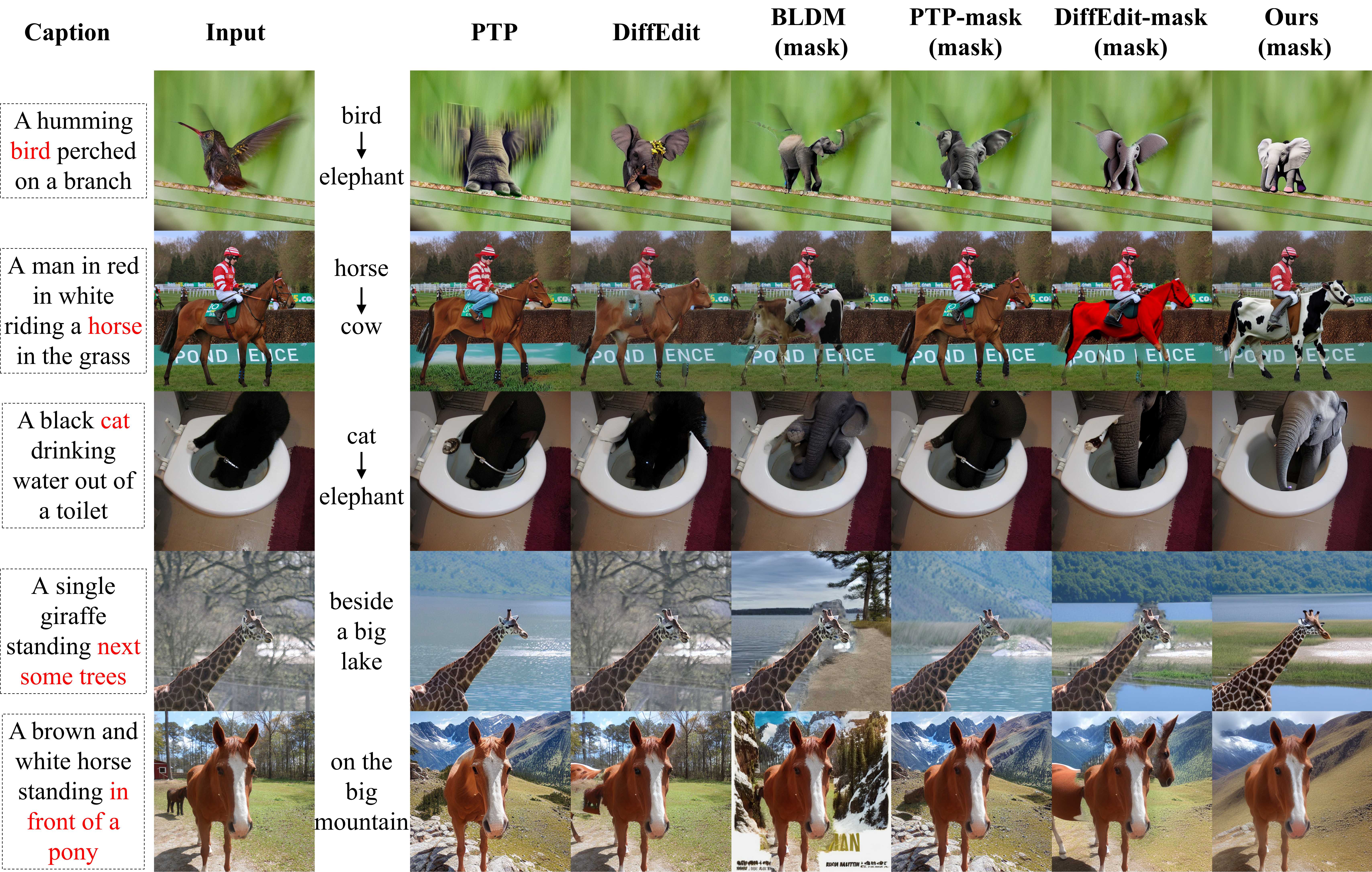}
	\caption{\textcolor{black}{Examples of edits on COCO \cite{coco} images. All the approaches adopt \texttt{Stable Diffusion v1-4} as the backbone, employing 50 steps of DDIM sampling. Better viewed online in color and zoomed in for details.}}
	\label{fig:coco}
\end{figure*}

\begin{figure*}[t]
	\centering
	\includegraphics[width=0.85\textwidth]{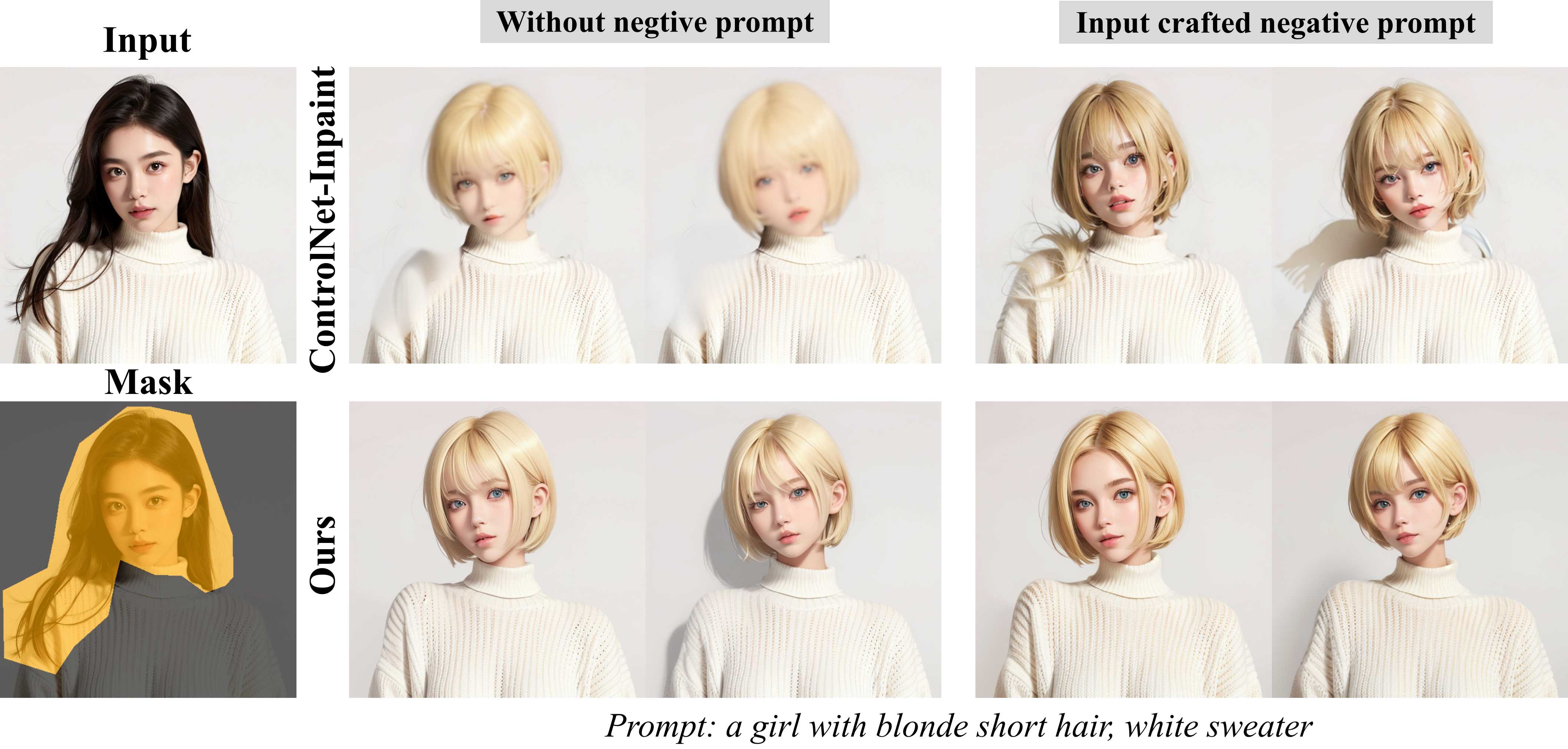}
	\caption{\textcolor{black}{Visual comparison of ControlNet-Inpaint \cite{controlnet} and our method in portrait editing.  The leftmost two columns display the editing results without negative prompts, whereas the rightmost two columns incorporate negative prompts. Notably, negative prompts significantly affect the outputs of ControlNet-Inpaint, and ControlNet-Inpaint often produces artifacts. In contrast, our method consistently delivers realistic edits irrespective of negative prompt usage.}}
	\label{fig:yellow-hair}
\end{figure*}

Similar observations can be found in real image editing. Figure \ref{fig:coco} shows the results on the COCOA-10k dataset. The complex scenes and text descriptions in the COCOA-10k dataset make it a good benchmark for evaluating generalization ability. Moreover, the masks in COCOA-10k are often incompatible with the target text prompts (e.g., the model is expected to generate a cat in an area shaped like a giraffe), allowing us to evaluate the robustness of mask-based methods to extreme masks. Even under such extreme masks, our method can still generate desired content while maintaining high quality and consistency. In contrast, other baselines either fail to generate desired objects/scenes or produce unwanted artifacts.

\begin{figure}[t]
	\centering
	\includegraphics[width=1\columnwidth]{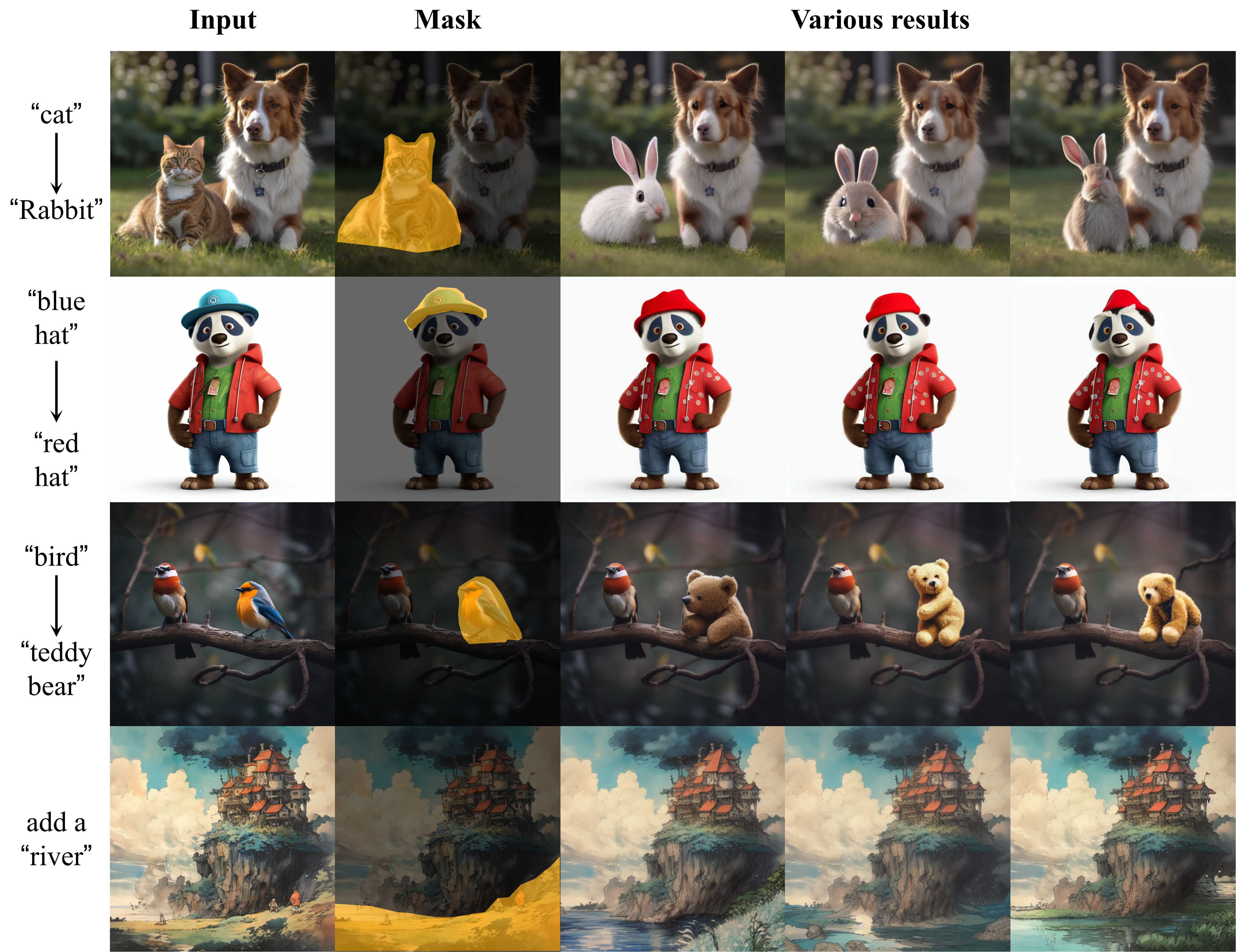}
	\caption{Our framework can synthesize realistic and diverse results according to the same source image and guided text prompt.}
	\label{fig:sup-various}
\end{figure}

Mask-based models can achieve satisfactory performance in background replacement by utilizing precise masks available in datasets like COCO \cite{coco}. However, such precise segmentation is not commonly available in real-world scenarios. In practice, users often provide rough masks, leading to foreground objects being segmented along with some original background information. For instance, when separating a horse from the grass, the extracted mask will inevitably include some portions of the surrounding grass. To assess the effectiveness of mask-based methods in practical settings, we provide dilated masks to these models. The dilated masks are visualized in the Appendix. As illustrated in Figure \ref{fig:coco}, while BLDM \cite{bldm} and DiffEdit-mask \cite{diffedit} successfully replace the background, there are remnants of the old background content along the boundaries, resulting in an unnatural fusion. In contrast, PFB-Diff effectively eliminates the interference of the old background associated with the mask, enabling a seamless fusion between the new scene and the foreground object. This demonstrates the ability of PFB-Diff to overcome challenges posed by imperfect masks and achieve superior results in terms of realistic background replacement.

\textcolor{black}{In Figure \ref{fig:yellow-hair}, we present a qualitative comparison between our method and ControlNet-Inpaint \cite{controlnet} for portrait editing. We use a fine-tuned model, the  \texttt{XXMix\_9realistic}\footnote{\url{https://civitai.com/models/47274/xxmix9realistic}} model, as the backbone of ControlNet-Inpaint and our approach. This model is fine-tuned on portrait images based on pre-trained \texttt{Stable Diffusion v1-5}\footnote{\url{https://huggingface.co/runwayml/stable-diffusion-v1-5}}. To fully utilize the generative capabilities of \texttt{XXMix\_9realistic}, we incorporate negative prompting \cite{negetive-prompt} as suggested by the authors of the model. The results demonstrate that in the absence of negative prompts, the outputs rendered by ControlNet-Inpaint exhibit a noticeable blurriness. Although the inclusion of negative prompts improves the performance of ControlNet-Inpaint, it still generates artifacts, such as unrealistic hair on the shoulder area (first row, fourth column) and implausible shadows in the background (first row, fifth column). In contrast, our method consistently produces satisfying editing outcomes, regardless of the provision of negative prompts, showcasing superior robustness. Furthermore, the shadows generated in the background by our method are more realistic. }

Among the evaluated methods, only PFB-Diff can simultaneously meet the requirements of high accuracy, high image quality, seamless fusion, and irrelevance preservation, in both object and background editing. Figure \ref{fig:sup-various} shows that the proposed method can generate a variety of plausible outcomes. More qualitative results can be found in the Appendix.

\begin{figure}[t]
	\centering
	\includegraphics[width=\columnwidth]{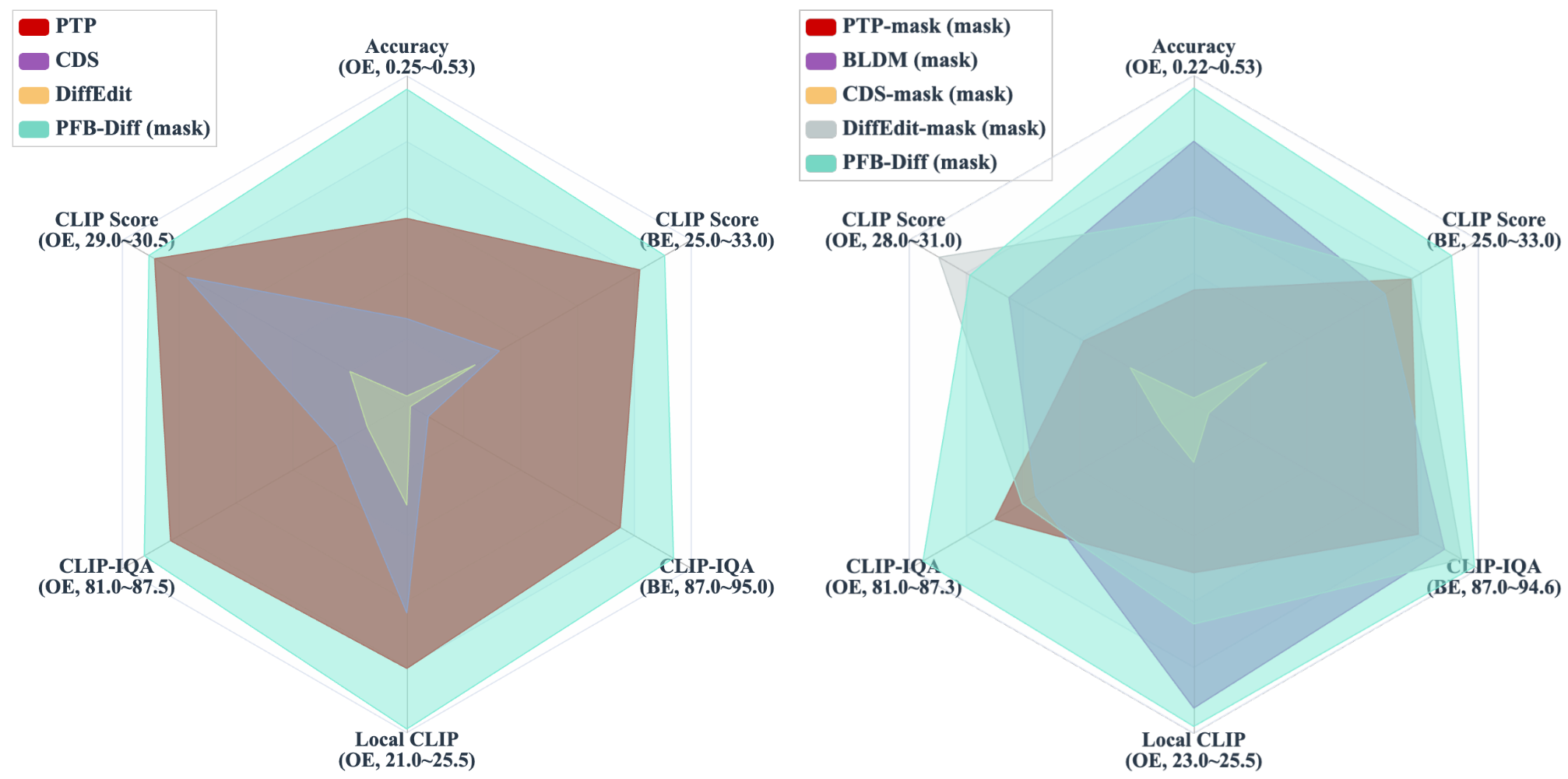}
	\caption{\redd{Quantitative comparison of various methods for object editing (OE) and background editing (BE) on 9,843 and 1,579 COCO \cite{coco} animal images, respectively. For ease of observation, we split the comparison of eight methods into two sub-figures, in which the results of PFB-Diff (mask) are reported twice. The term ``Accuracy (OE, 0.25$\sim$0.53)" refers to the accuracy calculated on the object editing (OE) task, with axis values ranging from 0.25 to 0.53. Proximity to the center of the radar chart indicates lower values, signifying inferior performance on the metric. Note that CLIP-IQA values are amplified 100 times. Refer to Section \ref{sec:set-up} for a detailed explanation of these metrics. All approaches, except CDS \cite{cds}, use \texttt{Stable Diffusion v1-4} with 50 DDIM sampling steps. CDS \cite{cds} also uses \texttt{Stable Diffusion v1-4} but with 200 inference steps. Better viewed online in color and zoomed in for details.}}
	\label{tab:compare}
\end{figure}

\begin{figure}[t]
	\centering
	\includegraphics[width=\columnwidth]{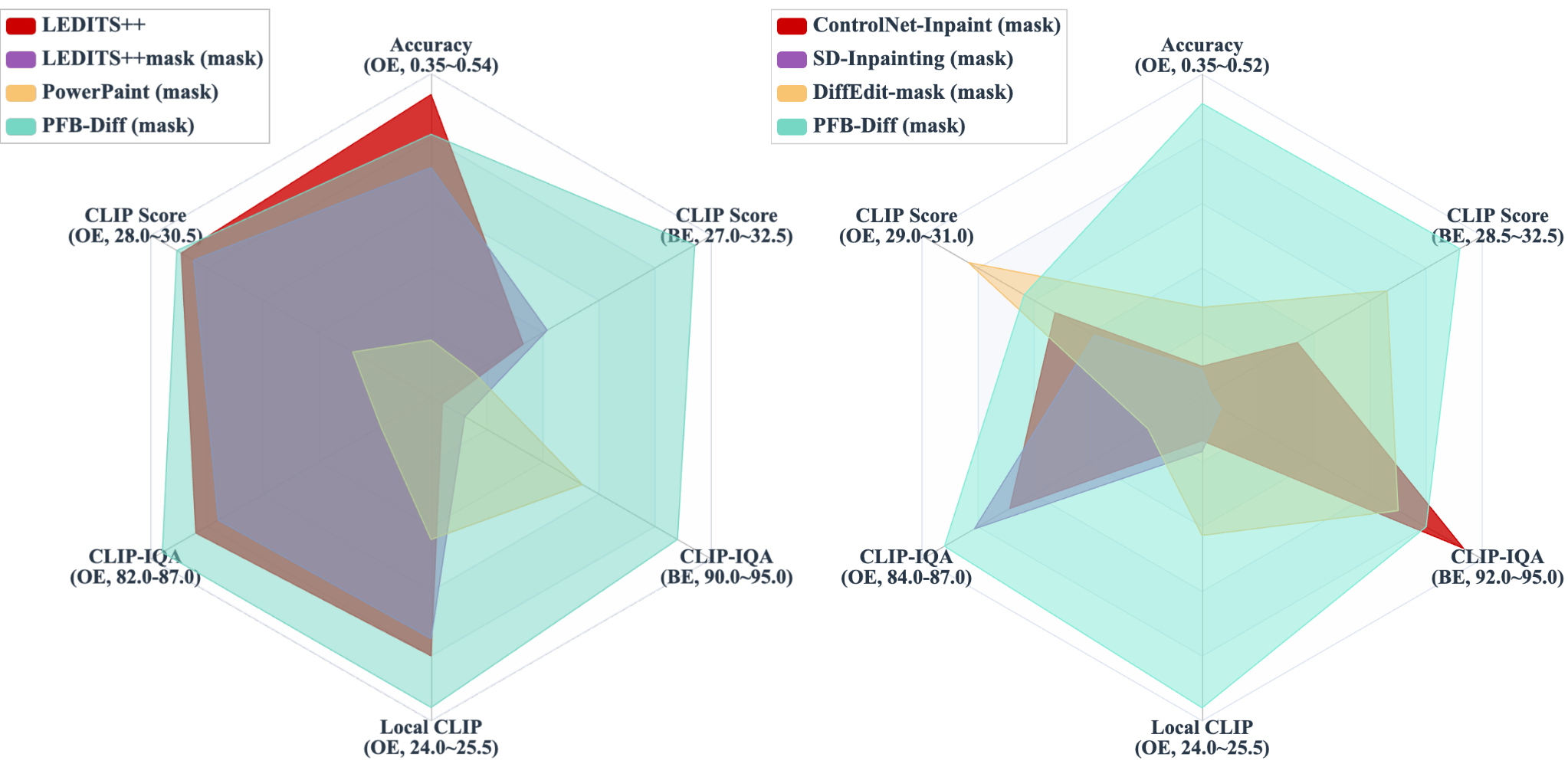}
	\caption{\redd{Quantitative comparison of various methods for object editing (OE) and background editing (BE) on 9,843 and 1,579 COCO \cite{coco} animal images, respectively. ControlNet-Inpaint \cite{controlnet}, DiffEdit-mask \cite{diffedit}, LEDITS++ \cite{ledits} and our method adopt \texttt{Stable Diffusion v1-5} as the backbone, while SD-Inpainting adopts \texttt{Stable Diffusion v1-5-inpainting} as the backbone. PowerPaint \cite{ppt} uses its fine-tuned model as the backbone. } }
	\label{tab:table-sd15}
\end{figure}

\subsection{Quantitative analysis}

\begin{table*}[t]
	\caption{\textcolor{black}{Quantitative results of different methods for object replacement and background replacement on 9,843 and 1,579 COCO \cite{coco} animal images, respectively. 
			The table reports four metrics: Accuracy (Acc), CLIP Score (CS), Local CLIP Score (LCS), and CLIP-IQA (CI). \redd{The comparative results reported are statistically significant, and the statistical demonstrations (p-values) are reported in the Appendix.} Detailed explanations of these metrics are provided in Section \ref{sec:set-up}. (Bold indicates the best result.)}} 
	\label{tab:coco-table}
	\begin{subtable}{\textwidth}
		\subcaption{The table provides the detailed results for Figure \ref{tab:compare}. \color{black}(The upper section lists mask-free methods, while the lower section includes mask-based methods.)}
		\centering
		\begin{tabular}{c|cccc|cc}
			\toprule
			\specialrule{0em}{2pt}{2pt}
			\multirow{2}{*}{Methods} &                            \multicolumn{4}{c|}{Object}                            &        \multicolumn{2}{c}{Background}         \\ \cline{2-7}
			\specialrule{0em}{2pt}{2pt}               &    Acc(\%) $\uparrow$     & CS $\uparrow$ & LCS $\uparrow$ &    CI $\uparrow$    & CS $\uparrow$ & CI $\uparrow$ \\ \midrule
			PTP \cite{ptp}                      &      40.85      &         30.33         &         24.62         &      0.864      &         31.55         & 0.930 \\
			DiffEdit \cite{diffedit}                 &      25.70      &         29.30         &         22.38         &      0.819      &         26.92              & 0.871 \\ 
			\color{black}CDS \cite{cds}                         &   32.33   &    	30.16      & 	23.86 &	 0.826 & 	27.60 &      0.876     \\ 	\midrule
			BLDM \cite{bldm}                     &      46.84      &         29.95         &         25.31         &      0.845      &        30.38         &  0.937\\
			PTP-mask \cite{ptp}                   &      32.80      &         29.16         &         24.28         &      0.854      &          31.11             & 0.930 \\
			DiffEdit-mask \cite{diffedit}              &      39.70      &    \textbf{30.69}     &         24.67         &      0.848      &          31.14             & 0.942 \\ 
			\redd{CDS-mask \cite{cds}}                          &  	22.60   &   	 28.67      & 	\text{23.44} & 	\text{0.817} & 	\text{27.05} &     	0.874     \\ 
			Ours                           & \textbf{ 51.87} &         30.36         &    \textbf{25.45}     & \textbf{ 0.870} &           \textbf{32.25}            & \textbf{0.945} \\ \bottomrule
			
		\end{tabular}
		\vspace{10pt}
	\end{subtable}
	\begin{subtable}{\textwidth}
		\centering
		\subcaption{The table provides the detailed results for Figure \ref{tab:table-sd15}.  \color{black}(The upper section lists mask-free methods, while the lower section includes mask-based methods.)}
		\begin{tabular}{c|cccc|cc}
			\toprule
			\specialrule{0em}{2pt}{2pt}
			\multirow{2}{*}{Methods} &                      \multicolumn{4}{c|}{Object}                      &  \multicolumn{2}{c}{Background}  \\ \cline{2-7}
			\specialrule{0em}{2pt}{2pt}               & Acc(\%) $\uparrow$ & CS $\uparrow$  & LCS $\uparrow$ & CI $\uparrow$  & CS $\uparrow$  &  CI $\uparrow$  \\ \midrule
			
			\redd{LEDITS++ \cite{ledits}   }              &   \textbf{52.81}   &     30.23      & \text{25.20} & \text{0.862} & \text{28.81} &      0.902    \\ \midrule
			ControlNet-Inpaint \cite{controlnet}           &       36.64        &     30.05      &     24.20      &     0.861      &     29.86      &     \textbf{ 0.948}       \\
			SD-Inpainting \cite{ldm}                 &       36.45        &     29.77      &     24.25      &     0.864      &     28.62      & 0.922 \\
			DiffEdit-mask \cite{diffedit}              &       39.74        & \textbf{30.67} &     24.64      &     0.846      &   31.14    &      0.941      \\
			
			\redd{LEDITS++mask \cite{ledits}   }                       &   \text{48.50}   &     30.12      & \text{25.12} & \text{0.858} & \text{29.28} &      0.906     \\
			\redd{PowerPaint \cite{ppt} }                         &   \text{38.36}   &     28.70      & \text{24.66} & \text{0.829} & \text{27.84} &      0.926    \\
			
			Ours                           &   50.46  &     30.27      & \textbf{25.44} & \textbf{0.868} & \textbf{32.18} &      0.944      \\ \bottomrule
		\end{tabular}
	\end{subtable}
	
\end{table*}

For quantitative comparison, we conduct object/background editings on 9,843 images of the COCOA-10k dataset. For PTP \cite{ptp}, DiffEdit \cite{diffedit}, PTP-mask, and DiffEdit-mask, the input data structure is a triplet consisting of an image, description, and target text prompt, while the target text prompt is replaced with a guided category (e.g., a dog) for BLDM \cite{bldm} and our method. For object editing, we replace words indicating animal categories in image descriptions with other categories to construct target text prompts. For background replacement, we select 1,597 images containing the word ``standing" from the COCOA-10k dataset and replace the context after ``standing" with random scenes, such as on the beach, in the snow, on the grass, or on a dusty road. Figure \ref{fig:coco} visualizes some examples of the COCOA-10k dataset.

\textcolor{black}{To intuitively compare the comprehensive performance of various methods across different task scenarios and metrics, we use radar charts to present the quantitative comparison results. As shown in Figure \ref{tab:compare} \& \ref{tab:table-sd15}, our method consistently outperforms others in various metrics, demonstrating its ability for precise editing while maintaining image quality. Despite DiffEdit-mask \cite{diffedit} achieving a slightly higher CLIP score in object editing, its local CLIP score and accuracy are notably lower than ours. This is due to the limitation of global CLIP score in accurately assessing the precision of local object edits, as discussed in Section \ref{sec:set-up}.}

\textcolor{black}{DiffEdit often struggles to make meaningful modifications to the original image, resulting in minimal changes and, consequently, the lowest accuracy and local CLIP score for object editing. Augmenting DiffEdit with user-provided masks significantly improves both accuracy in object editing and the CLIP score in background editing, emphasizing the shortcomings of self-predicted masks in DiffEdit. For PTP, utilizing user-provided masks does not enhance editing accuracy. These user-provided masks primarily serve to prevent unintended changes in unrelated areas during the editing process, as illustrated in Figure \ref{fig:exp-compare}.}

Detailed quantitative results for each method are provided in Table \ref{tab:coco-table}.

\subsection{User study}

\begin{table}[t]
	\begin{center}
		\caption{Human evaluation on semantic image editing. 1 is the worst, 4 is the best. Users rated ours as the best editing results.}  
		\label{tab:user-study}
		\begin{tabular}{ccccc}
			\toprule
			Method & PTP \cite{ptp} & BLDM \cite{bldm} & DiffEdit \cite{diffedit} & Ours   \\ \midrule
			Score $\uparrow$ &  2.10   &  2.47    &  1.72        &  \textbf{3.78}      \\ \bottomrule
		\end{tabular}
	\end{center}
\end{table}

To quantify overall human satisfaction, we conduct a user study on 28 participants. In the study, we use 30 groups of images, each group contains one input and four outputs. All these results in each group are presented side-by-side to participants. These 30 groups cover various editing tasks, including object replacement, background replacement, and object property editing. The input images include both real images and images generated by Midjourney. Participants are given unlimited time to rank the score from 1 to 4 (4 is the best, 1 is the worst) simultaneously considering the alignment with text prompts, realism, and faithfulness to the original input.
We report the average ranking score in Table \ref{tab:user-study}.  We observe that users prefer our results more than others.

\subsection{Ablation studies}

\begin{table}[t]
	\begin{center}
		\caption{Quantitative comparison of different variants of our method for object replacement and background replacement on 9,843 and 1,579 COCO \cite{coco} animal images, respectively. The table reports three metrics: Accuracy (Acc), CLIP Score (CS), and CLIP-IQA (CI). As shown, PFB-Diff achieves the best overall performance by leveraging all the proposed techniques.}  
		\label{tab:aba}
		
		\resizebox{\columnwidth}{!}{
		
		\begin{tabular}{c|ccc|cc}
			\toprule
			\specialrule{0em}{2pt}{2pt}
			\multirow{2}{*}{Methods} &                            \multicolumn{3}{c|}{Object}                            &        \multicolumn{2}{c}{Background}         \\ \cline{2-6}
			\specialrule{0em}{2pt}{2pt}           
			   &    Acc(\%) $\uparrow$     & CS $\uparrow$ &    CI $\uparrow$    & CS $\uparrow$ & CI $\uparrow$ \\ \midrule
			
			Baseline                &  40.15      &  29.88               & 0.846          &    31.30           & 0.943 \\
			+ PFB                   &  45.20        &   \textbf{30.38}            & 0.870           &        31.92            &  \textbf{0.947}   \\
			+ PFB, AM                            &  \textbf{51.87}            &  30.36      &          \textbf{0.870}            &  \textbf{32.25} & 0.945 \\ \bottomrule
			
		\end{tabular}}

	\end{center}
\end{table}
\begin{figure}[t]
	\centering
	\includegraphics[width=\columnwidth]{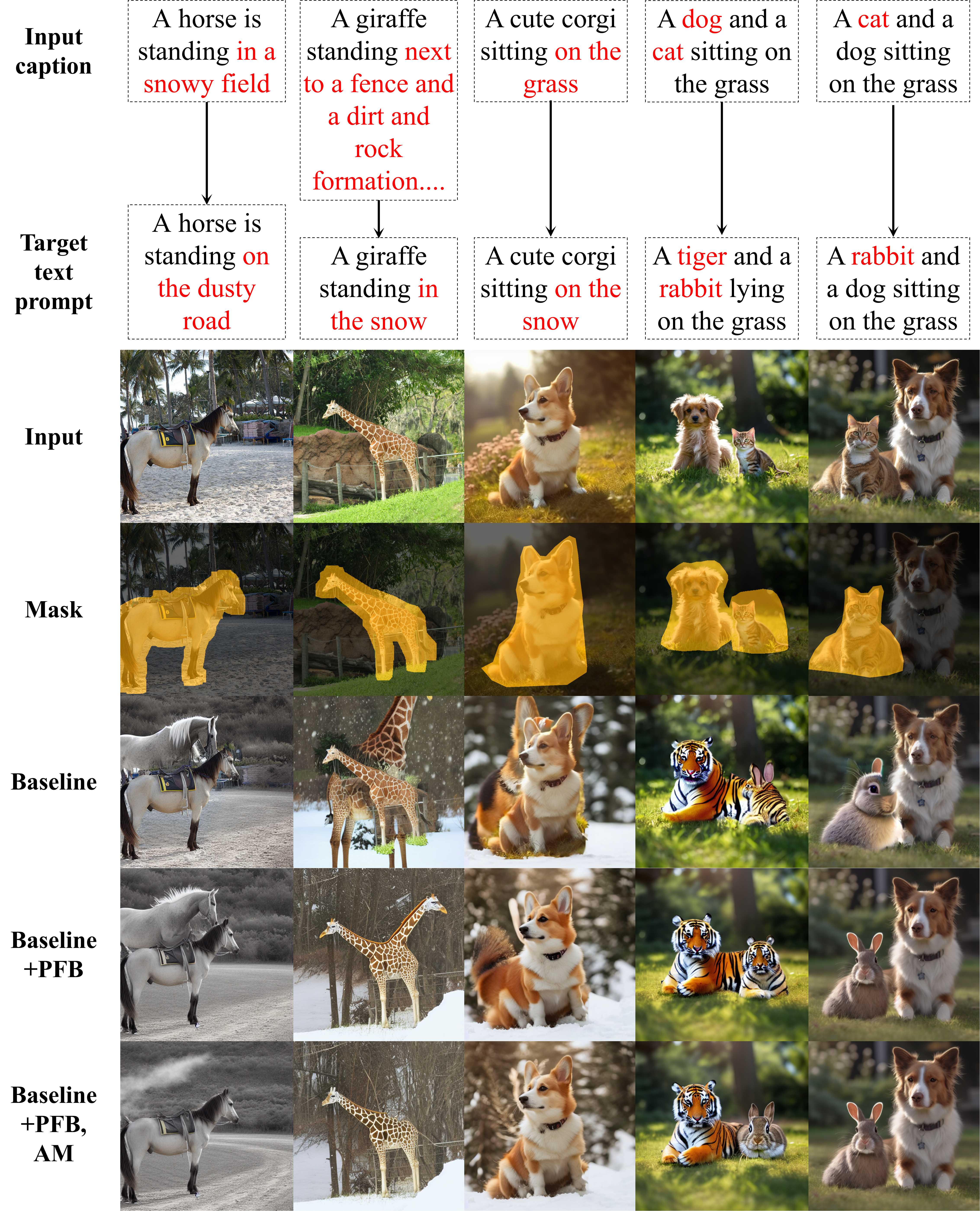}
	\caption{\red{Ablation study on the effects of the two modules of our method.}}
	\label{fig:aba}
\end{figure}

Our model consists of two pivotal components: the Progressive Feature Blending (PFB) technique and the Attention Masking (AM) mechanism.
In this section, we validate their effectiveness.  
To this end, we construct two variant models: 
1) Baseline, which adopts the latent-blending method utilized by previous works \cite{diffedit}. 
2) Baseline+PFB, which incorporates the PFB module. Baseline+PFB is equivalent to removing the attention masking mechanism from PFB-Diff.

Figure \ref{fig:aba} and Table \ref{tab:aba} display the qualitative and quantitative comparison results, respectively. As shown, PFB-Diff demonstrates the best overall performance by utilizing all the proposed techniques. The baseline often fails to generate the desired object or produce artifacts due to its latent-level blending, resulting in the worst quantitative results. By adding PFB to the baseline, both editing accuracy and CLIP score are improved. As shown in Figure \ref{fig:aba}, PFB can also facilitate a more seamless and natural blend between newly generated content and unrelated areas. \red{As shown in the fourth column of Figure~\ref{fig:aba}, the AM module plays a crucial role in multi-object replacement. It can prevent interference between different target objects by restricting their influence to specified regions, ensuring accurate generation of target objects in desired locations. In contrast, as illustrated in the fifth column of Figure~\ref{fig:aba}, the influence of the AM module is negligible in single object editing.} However, AM can improve the editing accuracy under extreme masks on the COCO \cite{coco} dataset. 

During background replacement, the results obtained from Baseline+PFB often exhibit additional objects around the foreground objects. As shown in the fourth row and first column of Figure \ref{fig:aba}, another horse appears around the original horse. When generating the new background, diffusion models are fed with the text prompt ``a horse is standing on a dusty road", but Baseline+PFB fails to restrict the generated horse to the position of the original horse. Masking the attention maps of specific words is introduced to address this issue. As shown in the last row of Figure \ref{fig:aba}, when we reintroduced AM into the model, we could observe an improvement in the aforementioned problems. 

\subsection{Generalization analysis}

\begin{figure}[t]
	\centering
	\includegraphics[width=\columnwidth]{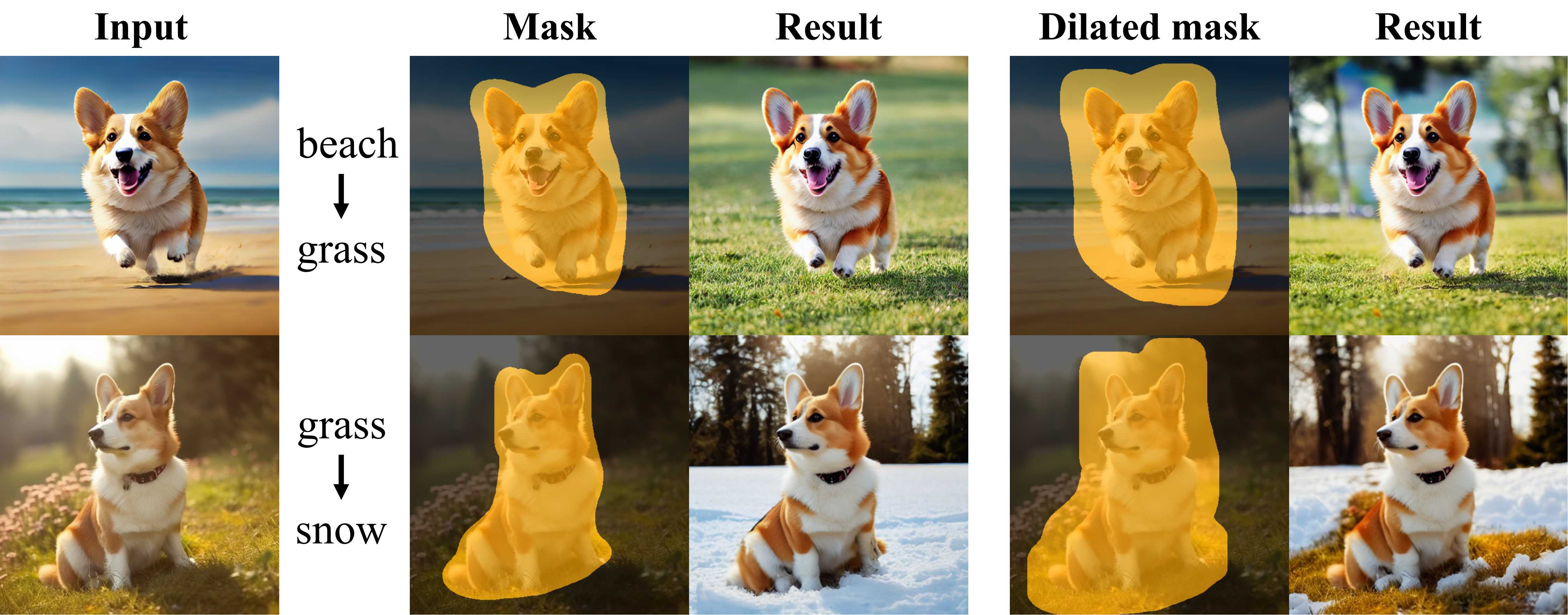}
	\caption{\red{The influence of mask coarseness on background editing.}}
	\label{fig:exp-general}
\end{figure}

\begin{figure*}[t]
	\centering
	\includegraphics[width=0.8\textwidth]{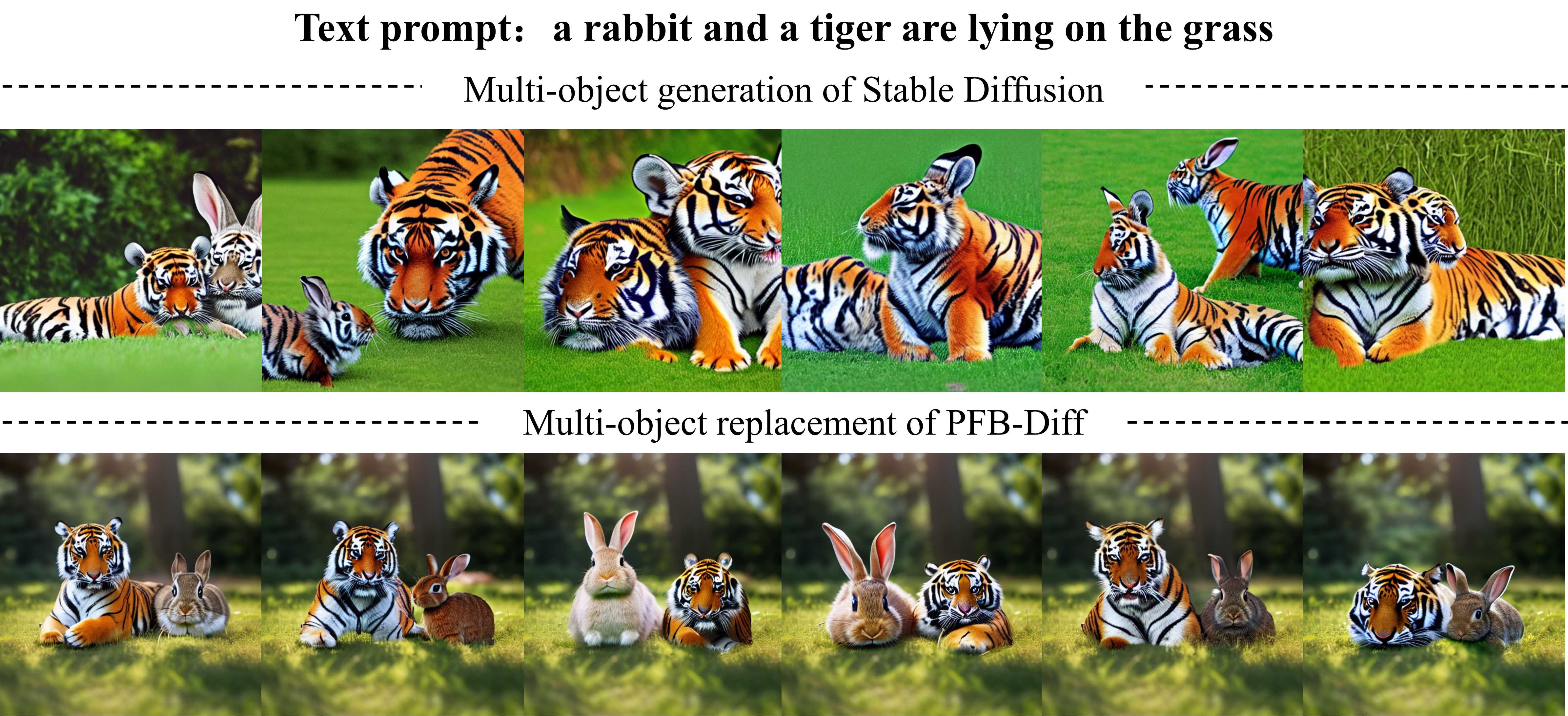}
	\caption{\red{The results of Stable Diffusion\cite{ldm} and PFB-Diff in multi-object generation.}}
	\label{fig:exp-general2}
\end{figure*}

\red{This section analyzes the generalization ability of PFB-Diff. Figure~\ref{fig:exp-general} shows the editing results of PFB-Diff when using masks of varying coarseness for background replacement.	The results indicate that when the mask covers only a small portion of the original background, PFB-Diff tends to ignore it. However, when the mask encompasses a substantial portion, PFB-Diff blends the original and new backgrounds naturally.  As shown in the last column of the second row of Figure~\ref{fig:exp-general}, PFB-Diff successfully transforms the original grass background into withered grass in the snow scene, blending naturally with the surrounding snow. This illustrates the robustness of PFB-Diff to masks of different roughness.}

\red{As shown in Figure~\ref{fig:exp-general2}, when the text prompt ``a tiger and a rabbit lying on the grass" is input into Stable Diffusion\cite{ldm}, the model often struggles to generate images that match the text description. Out of 40 random samples, only one successfully matched the description. In contrast, despite being based on the Stable Diffusion architecture, PFB-Diff achieves a high success rate of 37/40 in generating images that match the text description in object replacement scenarios, demonstrating its value in controlled generation applications.}

\section{Limitations and Discussions}
\begin{figure}[t]
	\centering
	\includegraphics[width=\columnwidth]{style-transfer.jpg}
	\caption{\red{The qualitative result of style transfer by Feature Renormalization}}
	\label{fig:style}
\end{figure}

\begin{figure}[t]
	\centering
	\includegraphics[width=\columnwidth]{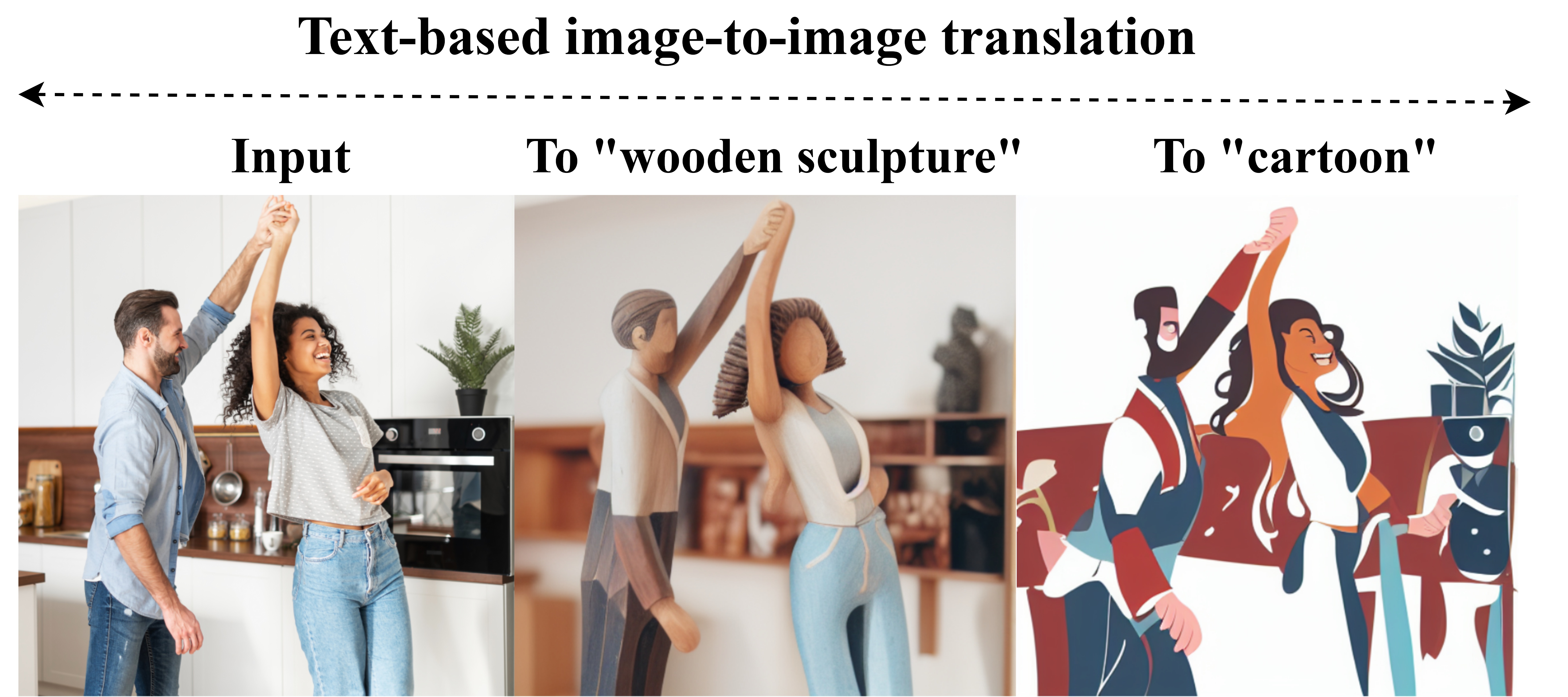}
	\caption{\red{The qualitative result of image-to-image translation by Wavelet based Feature Fusion.}}
	\label{fig:i2i}
\end{figure}
While our method has showcased promising results, it is still subject to a few limitations to be addressed in follow-up work. 
First of all, PFB-Diff requires users to provide rough masks for specifying regions of interest. While mask-based approaches often lead to more accurate editing results, they may pose an inconvenience in certain scenarios. For instance, if the mask of a small object is less than $8 \times 8$ in RGB space, it cannot be preserved after the naive downsample. \red{This problem arises from using the Stable Diffusion \cite{ldm} model as the backbone. To address this issue, we can retrain the Stable Diffusion model with a smaller downsampling factor and use a larger VAE in future research.}

Secondly, the size of the generated objects is determined by the size of the provided mask, which may result in unrealistic proportions. In Figure \ref{fig:exp-compare}, for example, the generated duck appears smaller than a flower. Nonetheless, this characteristic encourages users to create more imaginative compositions. \red{To control the size of the generated object, one straightforward idea is to incorporate words describing object size into text prompts. However, due to the limited capability of Stable Diffusion in language understanding, this approach may not be reliable in many situations. Another potential solution is to incorporate explicit geometric sizes into the generation process. In future work, we plan to introduce a trainable bounding box-image (bbox-image) cross-attention layer before the text-image cross-attention layer in each transformer block. This bbox-image layer will inject the bounding box into the image generation process, enabling the model to control the size of the generated objects.}

Finally, the feature-level manipulation presented in this paper is currently limited to mask-based blending. In future work, we plan to explore more complex feature-level operations to enable a wider range of image editing scenarios. \red{For example, we can combine the content and style of image features using feature renormalization techniques for example-based style transfer. Additionally, we plan to inject the layout of the target image into the generation process through wavelet-based feature fusion. The qualitative examples in Figures~\ref{fig:style} \& \ref{fig:i2i}  demonstrate the promising applications of these advanced feature-level manipulations.}

\begin{figure}[t]
	\centering
	\includegraphics[width=\columnwidth]{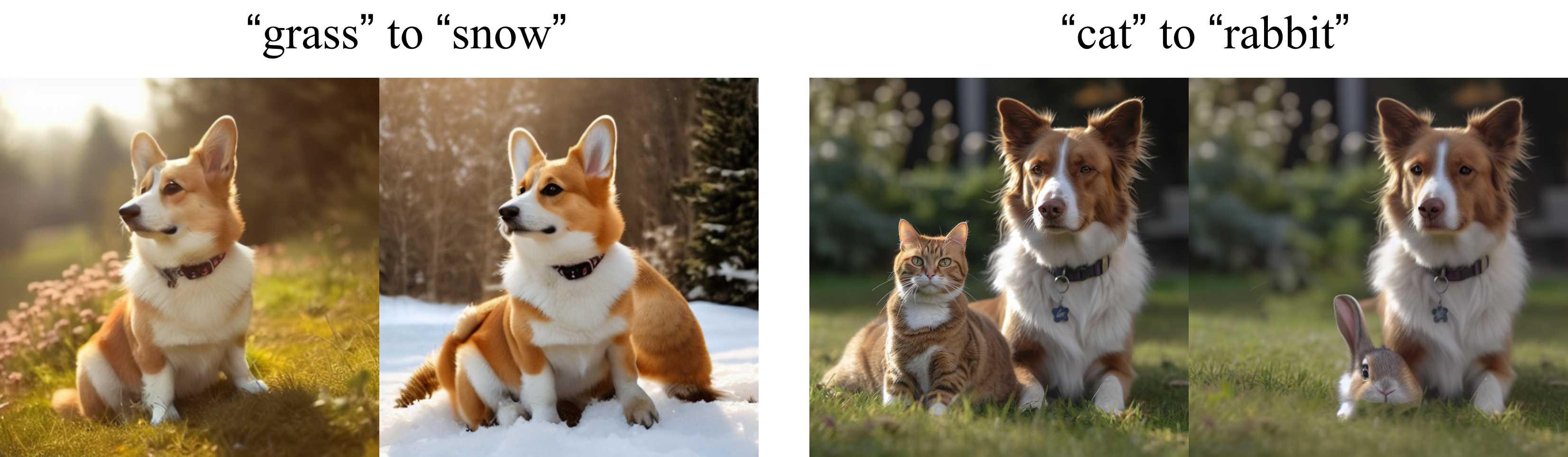}
	\caption{\red{Failure cases of PFB-Diff.}}
	\label{fig:sup-fail}
\end{figure}

\red{Figure \ref{fig:sup-fail} illustrates the failure cases of our method. While PFB-Diff generates images from random noise, it can be sensitive to initial noise. We fixed the target text prompt and the input image, then randomly sampled 20 results. As shown in Figure \ref{fig:sup-fail}, unwanted dog fur sometimes appears around the original dog during background replacement, and in one instance, the rabbit lacks a body. This limitation is inherited from Stable Diffusion \cite{ldm} and could potentially be addressed by using a text-to-image diffusion model that is more robust to the initial noise.}


\section{Conclusion}

This paper introduced PFB-Diff, a novel progressive feature blending method designed for diffusion-based image editing. By seamlessly blending irrelevant content with newly generated content across multiple levels of features, our approach achieves more natural and coherent editing results. Furthermore, we incorporate masked cross-attention maps to restrict the influence of specific words on the target area, resulting in enhanced performance.
Through empirical experiments on both real images and those generated by large-scale text-to-image models, we demonstrated the superior performance of our method compared to existing state-of-the-art techniques. Our approach effectively addresses various editing tasks, including object/background replacement and changing object properties, while preserving high-quality results and maintaining accurate image-text alignments.
Notably, our proposed method exhibits high efficiency, as it does not require fine-tuning or training. Both the PFB and AM modules are plug-and-play and can be easily adapted to any pre-trained text-to-image diffusion models. The above advantages position PFB-Diff as a promising direction for future research in text-driven image editing.

{\small
\bibliographystyle{ieee_fullname}
\bibliography{refs}

\begin{thebibliography}{10}\itemsep=-1pt

\bibitem{image2stylegan}
Rameen Abdal, Yipeng Qin, and Peter Wonka.
\newblock Image2stylegan: How to embed images into the stylegan latent space?
\newblock In {\em Proceedings of the IEEE international conference on computer
  vision}, pages 4432--4441, 2019.

\bibitem{image2stylegan++}
Rameen Abdal, Yipeng Qin, and Peter Wonka.
\newblock Image2stylegan++: How to edit the embedded images?
\newblock In {\em Proceedings of the IEEE/CVF Conference on Computer Vision and
  Pattern Recognition}, pages 8296--8305, 2020.

\bibitem{clip2stylegan}
Rameen Abdal, Peihao Zhu, John Femiani, Niloy Mitra, and Peter Wonka.
\newblock Clip2stylegan: Unsupervised extraction of stylegan edit directions.
\newblock In {\em ACM SIGGRAPH 2022 conference proceedings}, pages 1--9, 2022.

\bibitem{afzal2023visualization}
Shehzad Afzal, Sohaib Ghani, Mohamad~Mazen Hittawe, Sheikh~Faisal Rashid,
  Omar~M Knio, Markus Hadwiger, and Ibrahim Hoteit.
\newblock Visualization and visual analytics approaches for image and video
  datasets: A survey.
\newblock {\em ACM Transactions on Interactive Intelligent Systems},
  13(1):1--41, 2023.

\bibitem{bldm}
Omri Avrahami, Ohad Fried, and Dani Lischinski.
\newblock Blended latent diffusion.
\newblock {\em ACM Transactions on Graphics (TOG)}, 42(4):1--11, 2023.

\bibitem{bdm}
Omri Avrahami, Dani Lischinski, and Ohad Fried.
\newblock Blended diffusion for text-driven editing of natural images.
\newblock In {\em Proceedings of the IEEE/CVF Conference on Computer Vision and
  Pattern Recognition}, pages 18208--18218, 2022.

\bibitem{paint}
David Bau, Alex Andonian, Audrey Cui, YeonHwan Park, Ali Jahanian, Aude Oliva,
  and Antonio Torralba.
\newblock Paint by word.
\newblock {\em arXiv preprint arXiv:2103.10951}, 2021.

\bibitem{dalle3}
James Betker, Gabriel Goh, Li Jing, Tim Brooks, Jianfeng Wang, Linjie Li, Long
  Ouyang, Juntang Zhuang, Joyce Lee, Yufei Guo, et~al.
\newblock Improving image generation with better captions.
\newblock {\em Computer Science. https://cdn. openai. com/papers/dall-e-3.
  pdf}, 2:3, 2023.

\bibitem{ledits}
Manuel Brack, Felix Friedrich, Katharia Kornmeier, Linoy Tsaban, Patrick
  Schramowski, Kristian Kersting, and Apolin{\'a}rio Passos.
\newblock Ledits++: Limitless image editing using text-to-image models.
\newblock In {\em Proceedings of the IEEE/CVF Conference on Computer Vision and
  Pattern Recognition}, pages 8861--8870, 2024.

\bibitem{chen2023specref}
Songyan Chen and Jiancheng Huang.
\newblock Specref: A fast training-free baseline of specific
  reference-condition real image editing.
\newblock In {\em 2023 International Conference on Image Processing, Computer
  Vision and Machine Learning (ICICML)}, pages 369--375. IEEE, 2023.

\bibitem{ilvr}
Jooyoung Choi, Sungwon Kim, Yonghyun Jeong, Youngjune Gwon, and Sungroh Yoon.
\newblock Ilvr: Conditioning method for denoising diffusion probabilistic
  models.
\newblock {\em arXiv preprint arXiv:2108.02938}, 2021.

\bibitem{diffedit}
Guillaume Couairon, Jakob Verbeek, Holger Schwenk, and Matthieu Cord.
\newblock Diffedit: Diffusion-based semantic image editing with mask guidance.
\newblock In {\em The Eleventh International Conference on Learning
  Representations}, 2023.

\bibitem{guided-diff}
Prafulla Dhariwal and Alexander Nichol.
\newblock Diffusion models beat gans on image synthesis.
\newblock {\em Advances in Neural Information Processing Systems},
  34:8780--8794, 2021.

\bibitem{gen-1}
Patrick Esser, Johnathan Chiu, Parmida Atighehchian, Jonathan Granskog, and
  Anastasis Germanidis.
\newblock Structure and content-guided video synthesis with diffusion models.
\newblock In {\em Proceedings of the IEEE/CVF International Conference on
  Computer Vision}, pages 7346--7356, 2023.

\bibitem{textual-inversion}
Rinon Gal, Yuval Alaluf, Yuval Atzmon, Or Patashnik, Amit~Haim Bermano, Gal
  Chechik, and Daniel Cohen-or.
\newblock An image is worth one word: Personalizing text-to-image generation
  using textual inversion.
\newblock In {\em The Eleventh International Conference on Learning
  Representations}, 2022.

\bibitem{gal2022stylegan}
Rinon Gal, Or Patashnik, Haggai Maron, Amit~H Bermano, Gal Chechik, and Daniel
  Cohen-Or.
\newblock Stylegan-nada: Clip-guided domain adaptation of image generators.
\newblock {\em ACM Transactions on Graphics (TOG)}, 41(4):1--13, 2022.

\bibitem{emuvideo}
Rohit Girdhar, Mannat Singh, Andrew Brown, Quentin Duval, Samaneh Azadi,
  Sai~Saketh Rambhatla, Akbar Shah, Xi Yin, Devi Parikh, and Ishan Misra.
\newblock Emu video: Factorizing text-to-video generation by explicit image
  conditioning.
\newblock {\em arXiv preprint arXiv:2311.10709}, 2023.

\bibitem{gan}
Ian Goodfellow, Jean Pouget-Abadie, Mehdi Mirza, Bing Xu, David Warde-Farley,
  Sherjil Ozair, Aaron Courville, and Yoshua Bengio.
\newblock Generative adversarial nets.
\newblock {\em Advances in neural information processing systems}, 27, 2014.

\bibitem{mask-guided}
Shuyang Gu, Jianmin Bao, Hao Yang, Dong Chen, Fang Wen, and Lu Yuan.
\newblock Mask-guided portrait editing with conditional gans.
\newblock In {\em Proceedings of the IEEE Conference on Computer Vision and
  Pattern Recognition}, pages 3436--3445, 2019.

\bibitem{HARROU2022197}
Fouzi Harrou, Abdelhafid Zeroual, Mohamad~Mazen Hittawe, and Ying Sun.
\newblock Chapter 6 - recurrent and convolutional neural networks for traffic
  management.
\newblock In Fouzi Harrou, Abdelhafid Zeroual, Mohamad~Mazen Hittawe, and Ying
  Sun, editors, {\em Road Traffic Modeling and Management}, pages 197--246.
  Elsevier, 2022.

\bibitem{resnet}
Kaiming He, Xiangyu Zhang, Shaoqing Ren, and Jian Sun.
\newblock Deep residual learning for image recognition.
\newblock In {\em Proceedings of the IEEE conference on computer vision and
  pattern recognition}, pages 770--778, 2016.

\bibitem{ptp}
Amir Hertz, Ron Mokady, Jay Tenenbaum, Kfir Aberman, Yael Pritch, and Daniel
  Cohen-or.
\newblock Prompt-to-prompt image editing with cross-attention control.
\newblock In {\em The Eleventh International Conference on Learning
  Representations}, 2023.

\bibitem{hittawe2019abnormal}
Mohamad~Mazen Hittawe, Shehzad Afzal, Tahira Jamil, Hichem Snoussi, Ibrahim
  Hoteit, and Omar Knio.
\newblock Abnormal events detection using deep neural networks: application to
  extreme sea surface temperature detection in the red sea.
\newblock {\em Journal of Electronic Imaging}, 28(2):021012--021012, 2019.

\bibitem{hittawe2022efficient}
Mohamad~Mazen Hittawe, Sabique Langodan, Ouadi Beya, Ibrahim Hoteit, and Omar
  Knio.
\newblock Efficient sst prediction in the red sea using hybrid deep
  learning-based approach.
\newblock In {\em 2022 IEEE 20th International Conference on Industrial
  Informatics (INDIN)}, pages 107--117. IEEE, 2022.

\bibitem{ddpm}
Jonathan Ho, Ajay Jain, and Pieter Abbeel.
\newblock Denoising diffusion probabilistic models.
\newblock {\em Advances in Neural Information Processing Systems},
  33:6840--6851, 2020.

\bibitem{huang2024entwined}
Jiancheng Huang, Yifan Liu, Jiaxi Lv, and Shifeng Chen.
\newblock Entwined inversion: Tune-free inversion for real image faithful
  reconstruction and editing.
\newblock In {\em ICASSP 2024-2024 IEEE International Conference on Acoustics,
  Speech and Signal Processing (ICASSP)}, pages 2920--2924. IEEE, 2024.

\bibitem{huang2024sbcr}
Jiancheng Huang, Mingfu Yan, Yifan Liu, and Shifeng Chen.
\newblock Sbcr: Stochasticity beats content restriction problem in training and
  tuning free image editing.
\newblock In {\em Proceedings of the 2024 International Conference on
  Multimedia Retrieval}, pages 878--887, 2024.

\bibitem{huang2024wavedm}
Yi Huang, Jiancheng Huang, Jianzhuang Liu, Mingfu Yan, Yu Dong, Jiaxi Lyu,
  Chaoqi Chen, and Shifeng Chen.
\newblock Wavedm: Wavelet-based diffusion models for image restoration.
\newblock {\em IEEE Transactions on Multimedia}, 2024.

\bibitem{huang2024diffusion}
Yi Huang, Jiancheng Huang, Yifan Liu, Mingfu Yan, Jiaxi Lv, Jianzhuang Liu, Wei
  Xiong, He Zhang, Shifeng Chen, and Liangliang Cao.
\newblock Diffusion model-based image editing: A survey.
\newblock {\em arXiv preprint arXiv:2402.17525}, 2024.

\bibitem{ganspace}
Erik Härkönen, Aaron Hertzmann, Jaakko Lehtinen, and Sylvain Paris.
\newblock Ganspace: Discovering interpretable gan controls.
\newblock In {\em Proc. NeurIPS}, 2020.

\bibitem{stylegan}
Tero Karras, Samuli Laine, and Timo Aila.
\newblock A style-based generator architecture for generative adversarial
  networks.
\newblock In {\em Proceedings of the IEEE/CVF Conference on Computer Vision and
  Pattern Recognition}, pages 4401--4410, 2019.

\bibitem{stylegan2}
Tero Karras, Samuli Laine, Miika Aittala, Janne Hellsten, Jaakko Lehtinen, and
  Timo Aila.
\newblock Analyzing and improving the image quality of stylegan.
\newblock In {\em Proceedings of the IEEE/CVF Conference on Computer Vision and
  Pattern Recognition}, pages 8110--8119, 2020.

\bibitem{imagic}
Bahjat Kawar, Shiran Zada, Oran Lang, Omer Tov, Huiwen Chang, Tali Dekel, Inbar
  Mosseri, and Michal Irani.
\newblock Imagic: Text-based real image editing with diffusion models.
\newblock In {\em Proceedings of the IEEE/CVF Conference on Computer Vision and
  Pattern Recognition}, pages 6007--6017, 2023.

\bibitem{diffusionclip}
Gwanghyun Kim, Taesung Kwon, and Jong~Chul Ye.
\newblock Diffusionclip: Text-guided diffusion models for robust image
  manipulation.
\newblock In {\em Proceedings of the IEEE/CVF Conference on Computer Vision and
  Pattern Recognition}, pages 2426--2435, 2022.

\bibitem{sam}
Alexander Kirillov, Eric Mintun, Nikhila Ravi, Hanzi Mao, Chloe Rolland, Laura
  Gustafson, Tete Xiao, Spencer Whitehead, Alexander~C. Berg, Wan-Yen Lo, Piotr
  Dollar, and Ross Girshick.
\newblock Segment anything.
\newblock In {\em Proceedings of the IEEE/CVF International Conference on
  Computer Vision (ICCV)}, pages 4015--4026, October 2023.

\bibitem{yolov6}
Chuyi Li, Lulu Li, Yifei Geng, Hongliang Jiang, Meng Cheng, Bo Zhang, Zaidan
  Ke, Xiaoming Xu, and Xiangxiang Chu.
\newblock Yolov6 v3. 0: A full-scale reloading.
\newblock {\em arXiv preprint arXiv:2301.05586}, 2023.

\bibitem{coco}
Tsung-Yi Lin, Michael Maire, Serge Belongie, James Hays, Pietro Perona, Deva
  Ramanan, Piotr Doll{\'a}r, and C~Lawrence Zitnick.
\newblock Microsoft coco: Common objects in context.
\newblock In {\em Computer Vision--ECCV 2014: 13th European Conference, Zurich,
  Switzerland, September 6-12, 2014, Proceedings, Part V 13}, pages 740--755.
  Springer, 2014.

\bibitem{negetive-prompt}
Nan Liu, Shuang Li, Yilun Du, Antonio Torralba, and Joshua~B Tenenbaum.
\newblock Compositional visual generation with composable diffusion models.
\newblock In {\em European Conference on Computer Vision}, pages 423--439.
  Springer, 2022.

\bibitem{dpmsolver}
Cheng Lu, Yuhao Zhou, Fan Bao, Jianfei Chen, Chongxuan Li, and Jun Zhu.
\newblock {DPM}-solver: A fast {ODE} solver for diffusion probabilistic model
  sampling in around 10 steps.
\newblock In Alice~H. Oh, Alekh Agarwal, Danielle Belgrave, and Kyunghyun Cho,
  editors, {\em Advances in Neural Information Processing Systems}, 2022.

\bibitem{dpmsolver++}
Cheng Lu, Yuhao Zhou, Fan Bao, Jianfei Chen, Chongxuan Li, and Jun Zhu.
\newblock Dpm-solver++: Fast solver for guided sampling of diffusion
  probabilistic models.
\newblock {\em arXiv preprint arXiv:2211.01095}, 2022.

\bibitem{repaint}
Andreas Lugmayr, Martin Danelljan, Andres Romero, Fisher Yu, Radu Timofte, and
  Luc Van~Gool.
\newblock Repaint: Inpainting using denoising diffusion probabilistic models.
\newblock In {\em Proceedings of the IEEE/CVF Conference on Computer Vision and
  Pattern Recognition}, pages 11461--11471, 2022.

\bibitem{lv2024gpt4motion}
Jiaxi Lv, Yi Huang, Mingfu Yan, Jiancheng Huang, Jianzhuang Liu, Yifan Liu,
  Yafei Wen, Xiaoxin Chen, and Shifeng Chen.
\newblock Gpt4motion: Scripting physical motions in text-to-video generation
  via blender-oriented gpt planning.
\newblock In {\em Proceedings of the IEEE/CVF Conference on Computer Vision and
  Pattern Recognition}, pages 1430--1440, 2024.

\bibitem{sdedit}
Chenlin Meng, Yutong He, Yang Song, Jiaming Song, Jiajun Wu, Jun-Yan Zhu, and
  Stefano Ermon.
\newblock Sdedit: Guided image synthesis and editing with stochastic
  differential equations.
\newblock In {\em International Conference on Learning Representations}, 2021.

\bibitem{null-text}
Ron Mokady, Amir Hertz, Kfir Aberman, Yael Pritch, and Daniel Cohen-Or.
\newblock Null-text inversion for editing real images using guided diffusion
  models.
\newblock In {\em Proceedings of the IEEE/CVF Conference on Computer Vision and
  Pattern Recognition}, pages 6038--6047, 2023.

\bibitem{cds}
Hyelin Nam, Gihyun Kwon, Geon~Yeong Park, and Jong~Chul Ye.
\newblock Contrastive denoising score for text-guided latent diffusion image
  editing.
\newblock In {\em Proceedings of the IEEE/CVF Conference on Computer Vision and
  Pattern Recognition}, pages 9192--9201, 2024.

\bibitem{glide}
Alexander~Quinn Nichol, Prafulla Dhariwal, Aditya Ramesh, Pranav Shyam, Pamela
  Mishkin, Bob Mcgrew, Ilya Sutskever, and Mark Chen.
\newblock Glide: Towards photorealistic image generation and editing with
  text-guided diffusion models.
\newblock In {\em International Conference on Machine Learning}, pages
  16784--16804. PMLR, 2022.

\bibitem{pix2pix}
Gaurav Parmar, Krishna Kumar~Singh, Richard Zhang, Yijun Li, Jingwan Lu, and
  Jun-Yan Zhu.
\newblock Zero-shot image-to-image translation.
\newblock In {\em ACM SIGGRAPH 2023 Conference Proceedings}, pages 1--11, 2023.

\bibitem{ganimation}
Albert Pumarola, Antonio Agudo, Aleix~M Martinez, Alberto Sanfeliu, and
  Francesc Moreno-Noguer.
\newblock Ganimation: Anatomically-aware facial animation from a single image.
\newblock In {\em Proceedings of the European conference on computer vision
  (ECCV)}, pages 818--833, 2018.

\bibitem{fatezero}
Chenyang Qi, Xiaodong Cun, Yong Zhang, Chenyang Lei, Xintao Wang, Ying Shan,
  and Qifeng Chen.
\newblock Fatezero: Fusing attentions for zero-shot text-based video editing.
\newblock {\em arXiv:2303.09535}, 2023.

\bibitem{clip}
Alec Radford, Jong~Wook Kim, Chris Hallacy, Aditya Ramesh, Gabriel Goh,
  Sandhini Agarwal, Girish Sastry, Amanda Askell, Pamela Mishkin, Jack Clark,
  et~al.
\newblock Learning transferable visual models from natural language
  supervision.
\newblock In {\em International conference on machine learning}, pages
  8748--8763. PMLR, 2021.

\bibitem{pivotal}
Daniel Roich, Ron Mokady, Amit~H Bermano, and Daniel Cohen-Or.
\newblock Pivotal tuning for latent-based editing of real images.
\newblock {\em ACM Transactions on graphics (TOG)}, 42(1):1--13, 2022.

\bibitem{ldm}
Robin Rombach, Andreas Blattmann, Dominik Lorenz, Patrick Esser, and Bj{\"o}rn
  Ommer.
\newblock High-resolution image synthesis with latent diffusion models.
\newblock In {\em Proceedings of the IEEE/CVF Conference on Computer Vision and
  Pattern Recognition}, pages 10684--10695, 2022.

\bibitem{unet}
Olaf Ronneberger, Philipp Fischer, and Thomas Brox.
\newblock U-net: Convolutional networks for biomedical image segmentation.
\newblock In {\em Medical Image Computing and Computer-Assisted
  Intervention--MICCAI 2015: 18th International Conference, Munich, Germany,
  October 5-9, 2015, Proceedings, Part III 18}, pages 234--241. Springer, 2015.

\bibitem{dreambooth}
Nataniel Ruiz, Yuanzhen Li, Varun Jampani, Yael Pritch, Michael Rubinstein, and
  Kfir Aberman.
\newblock Dreambooth: Fine tuning text-to-image diffusion models for
  subject-driven generation.
\newblock In {\em Proceedings of the IEEE/CVF Conference on Computer Vision and
  Pattern Recognition}, pages 22500--22510, 2023.

\bibitem{imagen}
Chitwan Saharia, William Chan, Saurabh Saxena, Lala Li, Jay Whang, Emily~L
  Denton, Kamyar Ghasemipour, Raphael Gontijo~Lopes, Burcu Karagol~Ayan, Tim
  Salimans, et~al.
\newblock Photorealistic text-to-image diffusion models with deep language
  understanding.
\newblock {\em Advances in Neural Information Processing Systems},
  35:36479--36494, 2022.

\bibitem{schuhmannlaion}
Christoph Schuhmann, Romain Beaumont, Richard Vencu, Cade~W Gordon, Ross
  Wightman, Mehdi Cherti, Theo Coombes, Aarush Katta, Clayton Mullis, Mitchell
  Wortsman, et~al.
\newblock Laion-5b: An open large-scale dataset for training next generation
  image-text models.
\newblock In {\em Thirty-sixth Conference on Neural Information Processing
  Systems Datasets and Benchmarks Track}.

\bibitem{interfacegan}
Yujun Shen, Jinjin Gu, Xiaoou Tang, and Bolei Zhou.
\newblock Interpreting the latent space of gans for semantic face editing.
\newblock In {\em Proceedings of the IEEE/CVF Conference on Computer Vision and
  Pattern Recognition}, pages 9243--9252, 2020.

\bibitem{sohl2015deep}
Jascha Sohl-Dickstein, Eric Weiss, Niru Maheswaranathan, and Surya Ganguli.
\newblock Deep unsupervised learning using nonequilibrium thermodynamics.
\newblock In {\em International Conference on Machine Learning}, pages
  2256--2265. PMLR, 2015.

\bibitem{ddim}
Jiaming Song, Chenlin Meng, and Stefano Ermon.
\newblock Denoising diffusion implicit models.
\newblock In {\em International Conference on Learning Representations}, 2021.

\bibitem{improved-ddpm}
Yang Song and Stefano Ermon.
\newblock Improved techniques for training score-based generative models.
\newblock {\em Advances in neural information processing systems},
  33:12438--12448, 2020.

\bibitem{plug-and-play}
Narek Tumanyan, Michal Geyer, Shai Bagon, and Tali Dekel.
\newblock Plug-and-play diffusion features for text-driven image-to-image
  translation.
\newblock In {\em Proceedings of the IEEE/CVF Conference on Computer Vision and
  Pattern Recognition (CVPR)}, pages 1921--1930, June 2023.

\bibitem{attn}
Ashish Vaswani, Noam Shazeer, Niki Parmar, Jakob Uszkoreit, Llion Jones,
  Aidan~N Gomez, {\L}ukasz Kaiser, and Illia Polosukhin.
\newblock Attention is all you need.
\newblock {\em Advances in neural information processing systems}, 30, 2017.

\bibitem{clip-iqa}
Jianyi Wang, Kelvin~CK Chan, and Chen~Change Loy.
\newblock Exploring clip for assessing the look and feel of images.
\newblock In {\em AAAI}, 2023.

\bibitem{yang2021semantic}
Ceyuan Yang, Yujun Shen, and Bolei Zhou.
\newblock Semantic hierarchy emerges in deep generative representations for
  scene synthesis.
\newblock {\em International Journal of Computer Vision}, pages 1--16, 2021.

\bibitem{controlnet}
Lvmin Zhang, Anyi Rao, and Maneesh Agrawala.
\newblock Adding conditional control to text-to-image diffusion models.
\newblock In {\em Proceedings of the IEEE/CVF International Conference on
  Computer Vision (ICCV)}, pages 3836--3847, October 2023.

\bibitem{sine}
Zhixing Zhang, Ligong Han, Arnab Ghosh, Dimitris~N Metaxas, and Jian Ren.
\newblock Sine: Single image editing with text-to-image diffusion models.
\newblock In {\em Proceedings of the IEEE/CVF Conference on Computer Vision and
  Pattern Recognition}, pages 6027--6037, 2023.

\bibitem{indomain}
Jiapeng Zhu, Yujun Shen, Deli Zhao, and Bolei Zhou.
\newblock In-domain gan inversion for real image editing.
\newblock In {\em European Conference on Computer Vision}, pages 592--608.
  Springer, 2020.

\bibitem{ppt}
Junhao Zhuang, Yanhong Zeng, Wenran Liu, Chun Yuan, and Kai Chen.
\newblock A task is worth one word: Learning with task prompts for high-quality
  versatile image inpainting.
\newblock {\em arXiv preprint arXiv:2312.03594}, 2023.

\end{thebibliography}
}

\newpage
\appendix

\section{Implementation details of the compared methods}   \label{sec:sup-implementation-other}
For Blended Latent Diffusion\footnote{\url{https://github.com/omriav/blended-latent-diffusion}} (shorted as BLDM) \cite{bldm}, Prompt-to-Prompt\footnote{\url{https://github.com/google/prompt-to-prompt}} \cite{ptp} (shorted as PTP),  \redd{Contrastive Denoising Score\footnote{\url{https://github.com/HyelinNAM/ContrastiveDenoisingScore}} \cite{cds} (shorted as CDS), LEDITS++ \footnote{\url{https://huggingface.co/spaces/editing-images/leditsplusplus}} \cite{ledits} and PowerPaint\footnote{\url{https://github.com/open-mmlab/PowerPaint}} \cite{ppt}}, we adopt their official implementations.

DiffEdit \cite{diffedit} does not have an official implementation, whereas Lu et al. \cite{dpmsolver} released a popular re-implementation\footnote{\url{https://github.com/LuChengTHU/dpm-solver}} of DiffEdit in their paper. To accelerate the sampling, they use the dpm-solver \cite{dpmsolver} sampling method in their implementation instead of DDIM \cite{ddim} sampling used in the original paper of DiffEdit. To make a full re-implementation of DiffEdit, we slightly modified Lu's re-implementation and used the DDIM sampling method mentioned in the original paper.

We also conduct a comparison with mask-based DiffEdit, abbreviated as DiffEdit-mask, where we replace their self-predicted masks with user-provided masks. This variant essentially degenerates DiffEdit to SDEdit \cite{sdedit} + DDIM encoding, which serves as our baseline. In simple terms, SDEdit \cite{sdedit} + DDIM encoding is equivalent to PFB-Diff without progressive feature blending and masked attention, and also equivalent to DiffEdit without mask inference.

\textcolor{black}{For ControlNet-Inpaint \cite{controlnet} and SD-Inpainting \cite{ldm}, we utilize their implementations from the popular Stable Diffusion web UI project\footnote{\url{https://github.com/AUTOMATIC1111/stable-diffusion-webui}}. On the COCOA-10k dataset,  which lacks negative prompts, both ControlNet-Inpaint and SD-Inpainting often result in blurry outcomes when using the DDIM \cite{ddim} sampler. Therefore, they both employ the ``DPM++ 2M Karras" \cite{dpmsolver++} sampler with 30 steps for COCOA-10k, which is the default setting for inpainting.} \textcolor{black}{For ControlNet-Inpaint, we adopt its publicly available pre-trained model \texttt{control\_v11p\_sd15\_inpaint.pth}\footnote{\url{https://huggingface.co/lllyasviel/ControlNet-v1-1/tree/main}}. }

\section{COCO-animals-10k dataset}  \label{sec:our-data}
To quantitatively evaluate the performance of text-based image editing, we create a dataset called COCO-animals-10k, abbreviated as COCOA-10k. Specifically, we collect 9,850 images from the COCO dataset that include objects from 9 specific classes: dogs, cats, sheep, cows, horses, birds, elephants, zebras, and giraffes. By appropriately modifying the captions of the images, this dataset can be used to evaluate object replacement and background editing. To assess object replacement, we substitute animal-related words in the image captions with words representing diverse animal categories. For background replacement, we select 1,597 images from the COCOA-10k dataset that include the term ``standing". Next, we replace the text descriptions following ``standing" with random scenes, such as being on a beach, in the snow, on grass, on a big mountain, on a dusty road, or in a desert area. Figure \ref{fig:sup-coco-fore} \& \ref{fig:sup-coco-back} visualize some examples of the COCOA-10k dataset. 

\section{More analysis of PFB-Diff}

\begin{figure*}[h]
	\centering
	\includegraphics[width=0.9\textwidth]{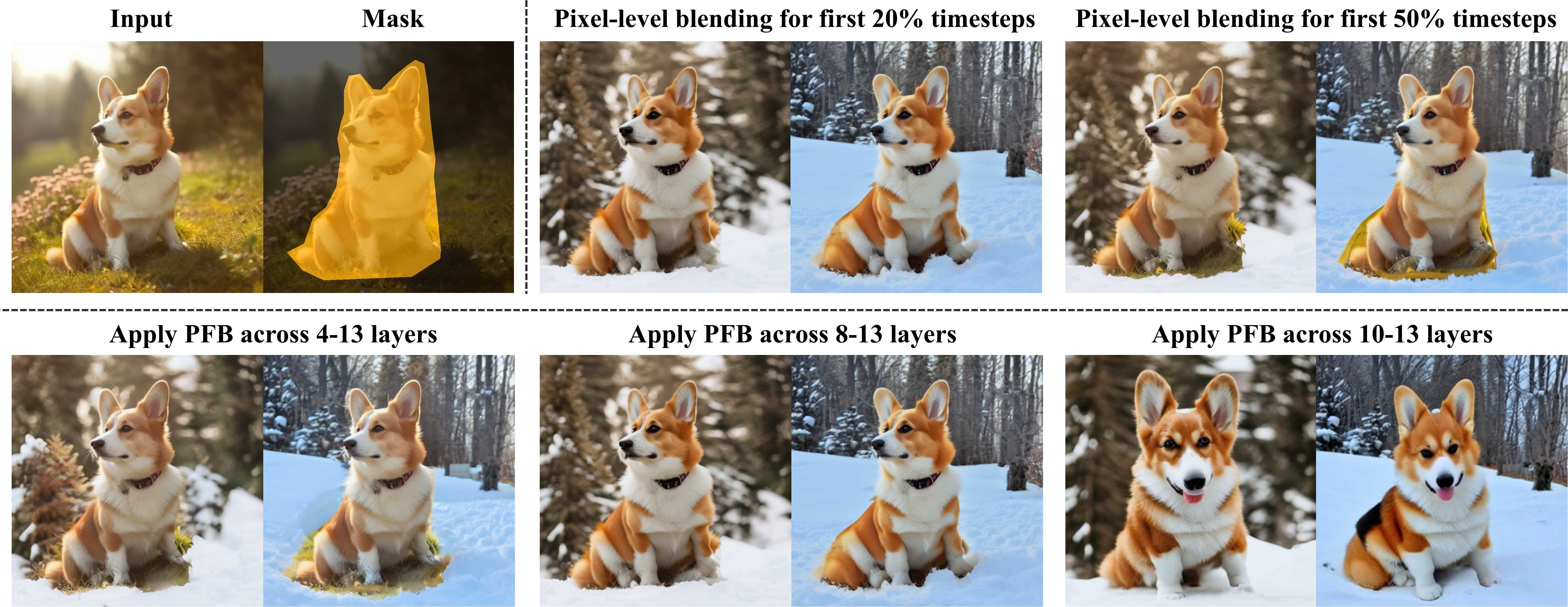}
	\caption{Illustration of the influence of some hyper-parameters in background editing. We randomly sample two initial Gaussian noises and fix them as the starting point for the denoising process under different hyper-parameter settings. Each pair showcases the editing results starting from these initial noises conditioned on ``a dog sitting on snow."}
	\label{fig:sup-hyper}
\end{figure*}

In Figure \ref{fig:sup-hyper}, we visualize the impact of timesteps we apply pixel-level blending and the number of layers we apply PFB. In general, applying PFB to the low-level feature map (layers 4-13) will keep too much redundant information, such as the information of the old background (e.g., the grass around the dog). However, if it is only applied to very high-level features (layers 10-13), the detailed information of the irrelevant area (the foreground dog) will be lost. Meanwhile, applying latent-level blending at more timesteps (50\% timesteps) will retain more details of the original image. As shown in Figure \ref{fig:sup-hyper}, applying PFB at layers 8-13 and applying latent-level blending for the first 20\% of timesteps is a good choice for background editing.

\section{\red{Statistical demonstrations of quantitative results}}
\red{We conduct two-sample t-tests on each metric, comparing our method with each  method, and report the results in Table~\ref{tab:coco-table-p}. The statistical analysis indicates that nearly all comparison results presented in Table 1 of the main paper are significant. Specifically, our method significantly outperforms others in terms of accuracy, local CLIP score, and CLIP-IQA for object replacement tasks, as well as in CLIP Score for background replacement tasks. }
\begin{table*}[t]
	
	\caption{\redd{Statistical demonstrations (p-values of two-sample t-tests) for the quantitative comparison in Table 1 of the main paper. ``a(b)" represents $a \times 10^{b}$. Values in red indicate that our method is not significantly better than the compared methods. ControlNet stands for ControlNet-Inpaint \cite{controlnet}.}} 
	\label{tab:coco-table-p}
	\begin{subtable}{\textwidth}
		\subcaption{The table provides the p-values for Table~1(a) of the main paper. }
		\centering
		\begin{tabular}{c|cccc|cc}
			\toprule
			\specialrule{0em}{2pt}{2pt}
			\multirow{2}{*}{Methods} &                  \multicolumn{4}{c|}{Object}                   & \multicolumn{2}{c}{Background}  \\ \cline{2-7}
			\specialrule{0em}{2pt}{2pt}               &    Accuracy     &    CS     &      LCS       &       CI        &       CS       &       CI       \\ \midrule
			PTP \cite{ptp}                          &    7.3(-55)    &   \reddd{0.51}    &    1.3(-29)      &     5.1(-3)      &   4.2(-10)    &    5.1(-7)      \\
			DiffEdit \cite{diffedit}                        &        0.00        & 1.4(-69) &     0.00      &      1.3(-101)      &   1.4(-320)   &     1.6(-68)      \\ 
			CDS  \cite{cds}                &     4.3(-189)     &  1.8(-4)    &   1.5(-95)   &   1.9(-80)    &   4.6(-269)    &  2.5(-66)   \\ \midrule
			BLDM \cite{bldm}                           &    2.1(-13)    & 1.1(-12) &    0.06      &      7.1(-31)      &   2.0(-54)    &     1.9(-3)      \\
			PTP-mask \cite{ptp}                         &   5.6(-175)    & 1.4(-92) &     1.0(-54)      &      6.1(-14)      &   1.0(-21)    &     2.5(-7)      \\
			DiffEdit-mask \cite{diffedit}                       &    5.2(-69)    &      \reddd{/}     &     2.6(-27)      &      1.7(-23)      &   4.0(-23)    &     \reddd{0.18}      \\ 
			CDS-mask \cite{cds}                &  0.00        &  2.1(-175)    &  6.3(-147)    &  4.6(-108)     &   1.8(-306)   &  6.5(-67)    \\ \bottomrule
		\end{tabular}
	\end{subtable}
	\tabcolsep0.06in 
	\begin{subtable}{\textwidth}
		\centering
		\subcaption{The table provides the p-values for Table 1(b) of the main paper.}
		\begin{tabular}{c|cccc|cc}
			\toprule
			\specialrule{0em}{2pt}{2pt}
			\multirow{2}{*}{Methods} &                      \multicolumn{4}{c|}{Object}                      &  \multicolumn{2}{c}{Background}  \\ \cline{2-7}
			\specialrule{0em}{2pt}{2pt}               & Accuracy  & CS  & LCS & CI   & CS   &  CI   \\ \midrule
			LEDITS++ \cite{ledits}                &      \reddd{/}    &   \reddd{0.62}   &  1.2(-3)    &    4.0(-3)   & 1.2(-179)     &  2.4(-32)    \\ \midrule
			ControlNet \cite{controlnet}               &      2.1(-87)      &   2.0(-4)    &    3.1(-58)    &     1.6(-3)      &   3.8(-80)    &   \reddd{/}        \\
			SD-Inpainting \cite{ldm}                     &      1.1(-94)      &   9.0(-17)   &    1.7(-52)    &    \reddd{0.15}   &   2.6(-166)   & 2.5(-11)  \\
			DiffEdit-mask \cite{diffedit}                &      9.1(-54)      &     \reddd{/}    &    4.2(-25)    &    4.3(-23)    &   6.3(-20)    &       \reddd{0.32}  \\
			LEDITS++mask \cite{ledits}                &   7.7(-3)      &   0.01    &   3.2(-5)   & 8.3(-5)      &   3.4(-108)   &  6.7(-26)    \\
			PowerPaint \cite{ppt}                &      1.9(-71)    &  2.5(-151)    &  3.6(-25)    &   2.3(-74)    & 1.9(-236)     &   3.9(-10)   \\
			\bottomrule
		\end{tabular}
	\end{subtable}
\end{table*}

\section{More qualitative results}

\begin{figure*}[h]
	\centering
	\includegraphics[width=\textwidth]{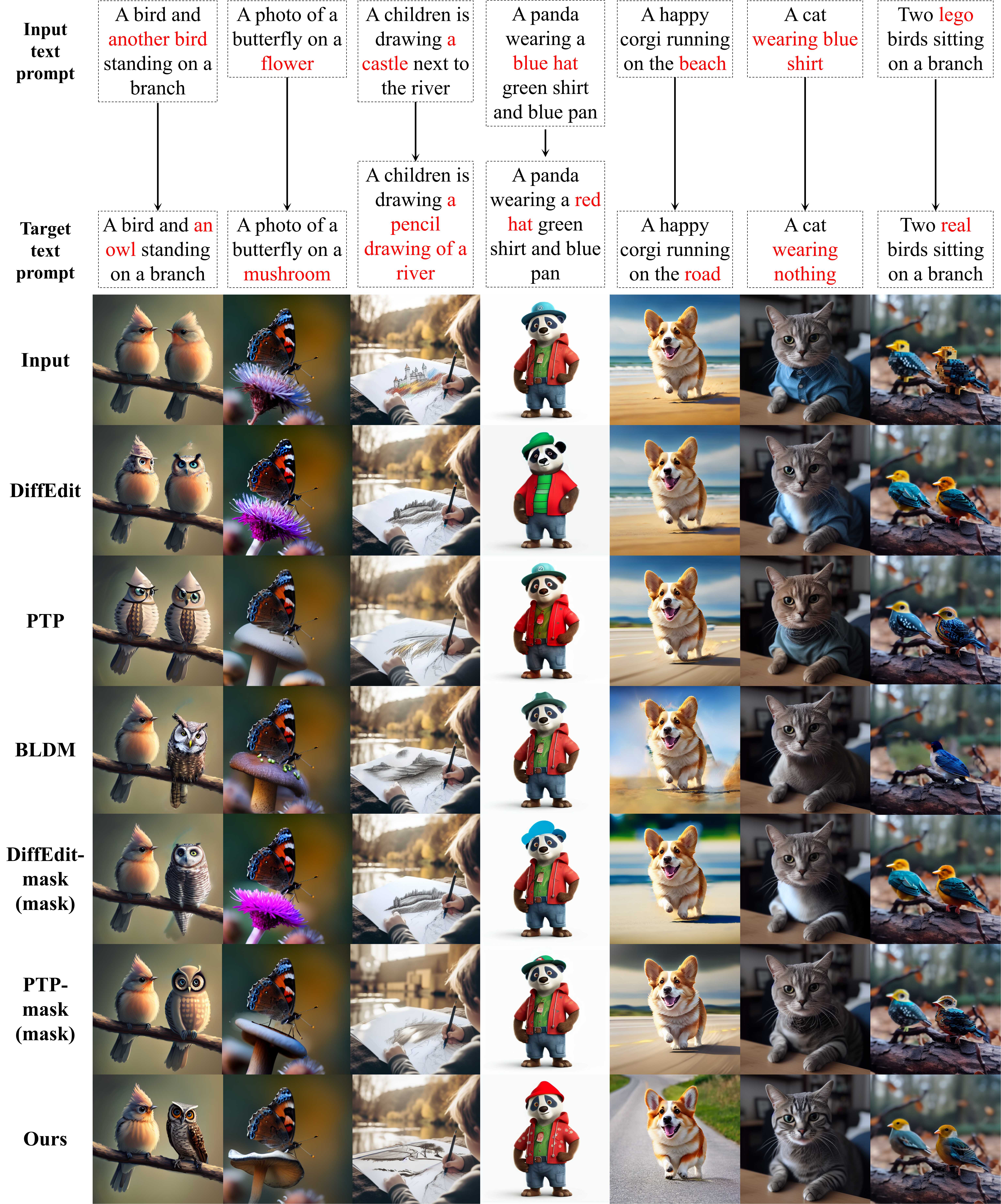}
	\caption{More qualitative comparisons on images generated by Midjourney. We compare our method with baselines on various editing tasks, including object replacement (columns 1-4), background editing (column 5), and object property editing (columns 6-7). For PTP \cite{ptp}, we tried our best to adjust the hyperparameters for each image to obtain the best result. Better viewed online in color and zoomed in for details.}
	\label{fig:sup-mid}
\end{figure*}

Figure \ref{fig:sup-mid} illustrates editing examples on images generated by Midjourney\footnote{\url{https://www.midjourney.com/home/?callbackUrl=\%2Fapp\%2F}}, in comparison with both mask-free editing methods, i.e., DiffEdit \cite{diffedit} and Prompt-to-Prompt \cite{ptp} (with Null-text Inversion \cite{null-text}), as well as mask-based editing methods, i.e., BLDM \cite{bldm}, DiffEdit-mask, and PTP-mask. PFB-Diff generally performs more targeted and accurate edits, leaving irrelevant regions intact. Consider for example the fourth column of Figure \ref{fig:sup-mid}, where PFB-Diff can accurately produce a red hat, while other methods fail to generate a hat of specified colors. 

\begin{figure*}[t]
	\centering
	\includegraphics[width=\textwidth]{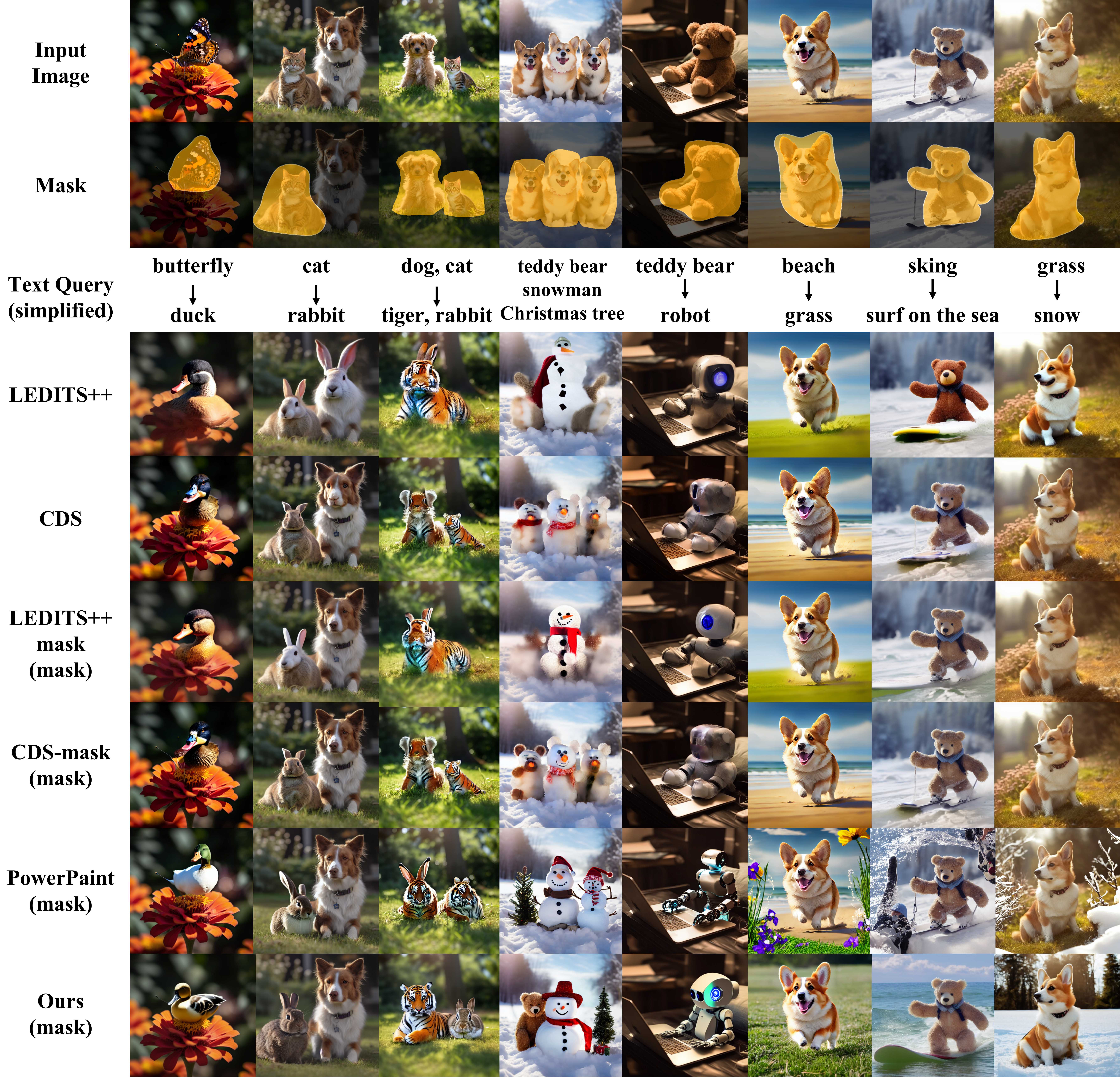}
	\caption{\redd{Examples of edits on images obtained from Midjourney. For mask-based methods marked with ``(mask)", we use the manually annotated rough labels shown in the second row. For LEDITS++ \cite{ledits}, we tried our best to adjust the target prompts and hyperparameters to obtain the best result. Better viewed online in color and zoomed in for details.} }
	\label{fig:sup-compare2}
\end{figure*}

\redd{Figure \ref{fig:sup-compare2} displays the editing results from Contrastive Denoising Score \cite{cds} (abbreviated as CDS), PowerPaint \cite{ppt}, and LEDITS++ \cite{ledits}. Notably, the existing methods face considerable challenges in background editing and multi-object replacement. When editing multiple objects simultaneously, there exists mutual interference between different objects in the existing methods, causing some not to be accurately generated.  With the proposed attention masking mechanism, PFB-Diff effectively avoids such interference and accurately generates the target objects. For instance, as shown in the third column of Figure \ref{fig:sup-compare2}, while other methods only produce a tiger with visible artifacts, PFB-Diff generates both a realistic tiger and a rabbit without such flaws.}

\begin{figure*}[h]
	\centering
	\includegraphics[width=\textwidth]{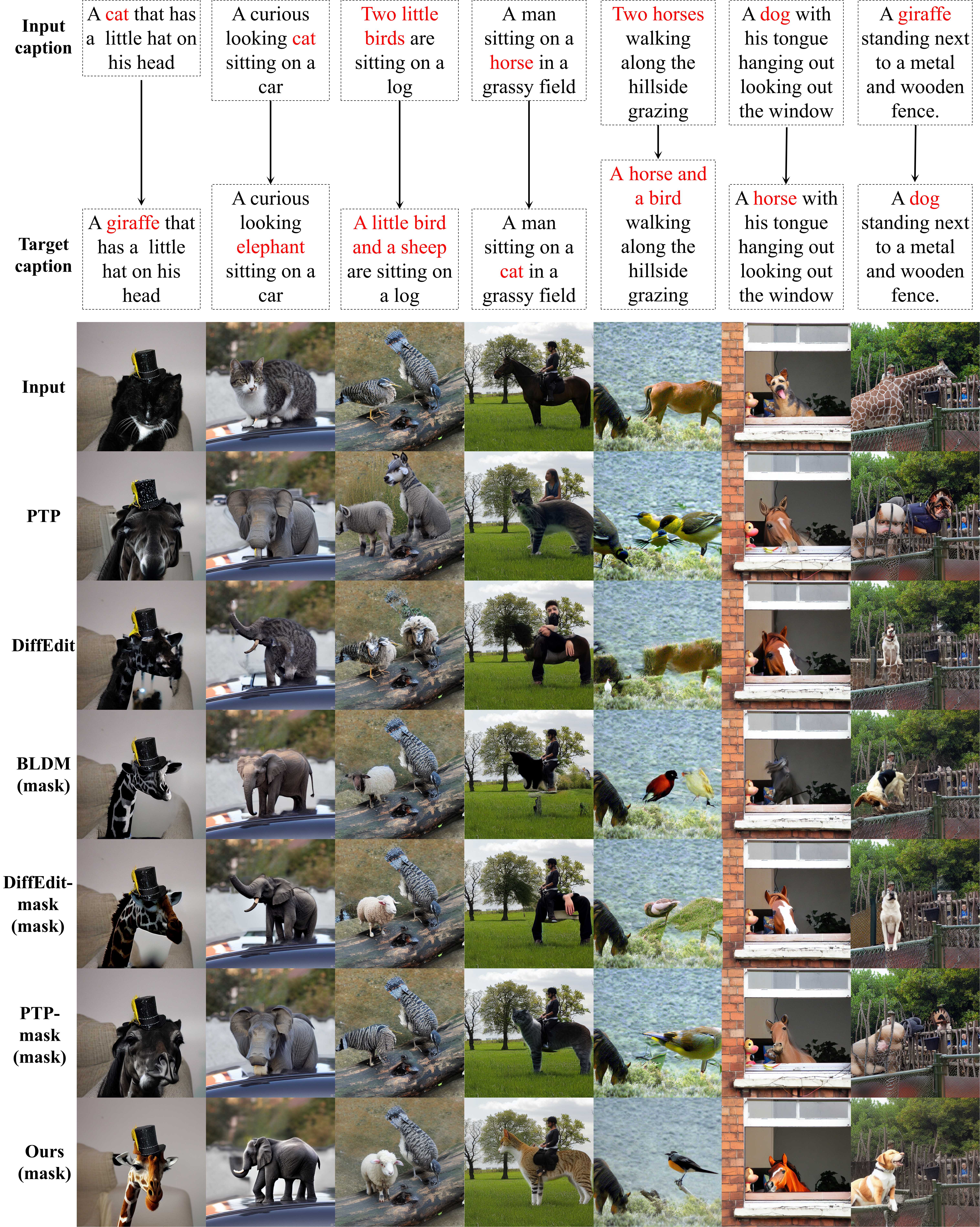}
	\caption{Qualitative results of different methods on object editing on COCO \cite{coco} images. Better viewed online in color and zoomed in for details.}
	\label{fig:sup-coco-fore}
\end{figure*}

\begin{figure*}[h]
	\centering
	\includegraphics[width=\textwidth]{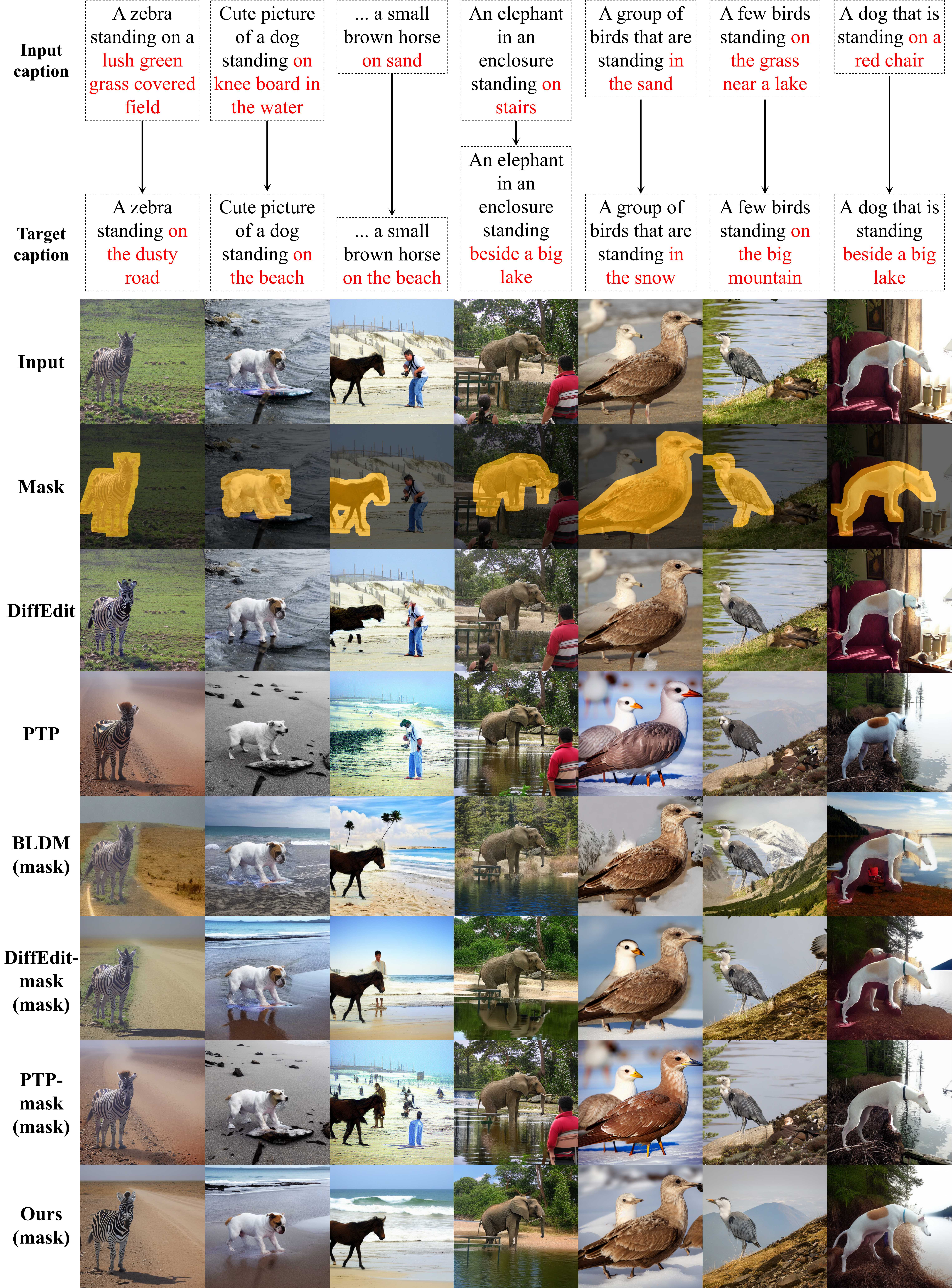}
	\caption{Qualitative results of different methods on background editing on COCO \cite{coco} images. The first row shows input images, and the second row visualizes the masks used in mask-based methods.}
	\label{fig:sup-coco-back}
\end{figure*}

Additional qualitative examples of editings on COCO \cite{coco} images are shown in Figure \ref{fig:sup-coco-fore} (object replacement) and Figure  \ref{fig:sup-coco-back} (background replacement). 

\section{Detailed comparison with closely related methods}
Plug-and-play \cite{plug-and-play} is a current method that utilizes the generative power of pre-trained diffusion models to perform text-driven image-to-image translation. It incorporates a feature injection approach that preserves the layout of the source image while translating it. Specifically,  plug-and-play \cite{plug-and-play} starts with a DDIM inversion \cite{ddim} of the source image, which then diverges into two distinct sampling pathways: one for reconstructing the source image and another for generating the target image. During the generation of the target image, features extracted from the reconstruction pathway are injected into the generation process. Although both plug-and-play and PFB-Diff employ feature manipulations, PFB-Diff differs from plug-and-play in three aspects.
Firstly, PFB-Diff is designed for high-quality localized editing, leveraging feature-level fusion to seamlessly integrate newly generated content into the source image,  emphasizing the rich semantics embedded in high-level features. Conversely,  plug-and-play focuses on maintaining the layout of the source image during translation by using feature injection, emphasizing spatial information embedded in features.
Secondly, in our approach, guided features are extracted during the DDIM inversion process, which preserves visual details in the non-target areas of the image. In contrast,  plug-and-play requires an additional sampling stage for extracting guided features.
Lastly, our method employs mask-based blending to fuse guided and target features. In contrast,  plug-and-play completely replaces the target features with the guided ones.

\end{document}